\documentclass[10pt,twocolumn,letterpaper]{article}

\usepackage{iccv}
\usepackage{times}
\usepackage{epsfig}
\usepackage{graphicx}
\usepackage{amsmath}
\usepackage{amssymb}

\usepackage{xcolor}
\usepackage{comment}
\usepackage{float}
\usepackage[inline]{enumitem}
\newcommand{\rvs}[1]{}

\usepackage{booktabs} % tables
\usepackage{multirow}
\usepackage{mathtools}
\usepackage[normalem]{ulem}
\usepackage{tabu}
\usepackage{subfigure}
\usepackage[ruled,linesnumbered]{algorithm2e}
\usepackage{xfrac}
\usepackage{caption}
\usepackage{color}
\usepackage{dblfloatfix}
\usepackage[symbol]{footmisc}
\usepackage{float}
\usepackage{bm}
\usepackage{afterpage}

\DeclareMathOperator*{\softargmax}{softargmax}
\newcommand{\round}[1]{\ensuremath{\lfloor#1\rceil}}

\usepackage{arydshln}

\makeatletter
\def\adl@drawiv#1#2#3{%
        \hskip.5\tabcolsep
        \xleaders#3{#2.5\@tempdimb #1{1}#2.5\@tempdimb}%
                #2\z@ plus1fil minus1fil\relax
        \hskip.5\tabcolsep}
\newcommand{\cdashlinelr}[1]{%
  \noalign{\vskip\aboverulesep
           \global\let\@dashdrawstore\adl@draw
           \global\let\adl@draw\adl@drawiv}
  \cdashline{#1}
  \noalign{\global\let\adl@draw\@dashdrawstore
           \vskip\belowrulesep}}
\makeatother

\newcommand{\beginsupplement}{%
        \setcounter{table}{0}
        \renewcommand{\thetable}{S\arabic{table}}%
        \setcounter{figure}{0}
        \renewcommand{\thefigure}{S\arabic{figure}}%
     }

\makeatletter
\newcommand\refwithdefault[2]{%
  \@ifundefined{r@#1}{%
    #2%
  }{%
    \ref{#1}%
  }%
}
\makeatother
\usepackage[pagebackref=true,breaklinks=true,letterpaper=true,colorlinks,bookmarks=false]{hyperref}

\iccvfinalcopy % *** Uncomment this line for the final submission

\def\iccvPaperID{6240} % *** Enter the ICCV Paper ID here

\ificcvfinal\fi

\begin{document}

%%%%%%%%% TITLE
\title{Physics-based Differentiable Depth Sensor Simulation}

% \author{Benjamin Planche$^1$\\
% $^1$Siemens Technology\\
% {\tt\small benjamin.planche@siemens.com}
% \and
% Rajat Vikram Singh$^{1*,2}$\\
% $^2$NVIDIA\\
% {\tt\small rajats@alumni.cmu.edu}
% }
% \author{Benjamin Planche$^1$\\
% Siemens Technology\\
% {\tt\small benjamin.planche@siemens.com}
% \and
% Rajat Vikram Singh\\
% Siemens Technology\\
% {\tt\small rajats@alumni.cmu.edu}
% }

\author{$\quad$Benjamin Planche $\quad\quad\quad\quad$ Rajat Vikram Singh$^{\ddag}$\\
Siemens Technology\\
{\tt\small benjamin.planche@siemens.com}, {\tt\small rajats@alumni.cmu.edu}
}

\maketitle

%%%%%%%%% ABSTRACT
\begin{abstract}
Gradient-based algorithms are crucial to modern com\-puter-\-vision and  graphics applications, enabling learning-based optimization and inverse problems. For example, photorealistic differentiable rendering pipelines for color images have been proven highly valuable to applications aiming to map 2D and 3D domains. 
However, to the best of our knowledge, no effort has been made so far towards extending these gradient-based methods to the generation of depth (2.5D) images, as simulating structured-light depth sensors implies solving complex light transport and stereo-matching problems.
In this paper, we introduce a novel end-to-end differentiable simulation pipeline for the generation of realistic 2.5D scans, built on physics-based 3D rendering and custom block-matching algorithms. Each module can be differentiated \wrt sensor and scene parameters; \eg, to automatically tune the simulation for new devices over some provided scans or to leverage the pipeline as a 3D-to-2.5D transformer within larger computer-vision applications. Applied to the training of deep-learning methods for various depth-based recognition tasks (classification, pose estimation, semantic segmentation), our simulation greatly improves the performance of the resulting models on real scans, thereby demonstrating the fidelity and value of its synthetic depth data compared to previous static simulations and learning-based domain adaptation schemes.

\end{abstract}

% \footnotetext[1]{Work done while at Siemens Technology.}
\footnotetext[3]{\scriptsize Now at NVIDIA.}

%%%%%%%%% BODY TEXT

\vspace{-.1em}

\section{Introduction}
Progress in computer vision has been dominated by deep neural networks trained over large amount of data, usually labeled. The deployment of these solutions into real-world applications is, however, often hindered by the cost (time, manpower, access, \etc) of capturing and annotating exhaustive training datasets of target objects or scenes. To partially or completely bypass this hard data requirement, an increasing number of solutions are relying on synthetic images rendered from 3D databases for their training~\cite{nips12:d3d,nips12:crdl,iccv13:fpe,
planche2017depthsynth,zakharov2019deceptionnet,planche2020bridging}, leveraging advances in computer graphics~\cite{schlick1994inexpensive,pharr2016physically}. Indeed, physics-based rendering methods are slowly but surely closing the visual gap between real and synthetic color image distributions, simulating complex optical phenomena (\eg, realistic light transport, lens aberrations, Bayer demosaicing, \etc). While these extensive tools still require domain knowledge to be properly parameterized for each new use-case (\wrt scene content, camera properties, \etc), their positive impact on the training of color-based visual recognition algorithms has been well documented already~\cite{denninger2019blenderproc,hodavn2019photorealistic}.

\begin{figure}
\centering
\includegraphics[width=\linewidth]{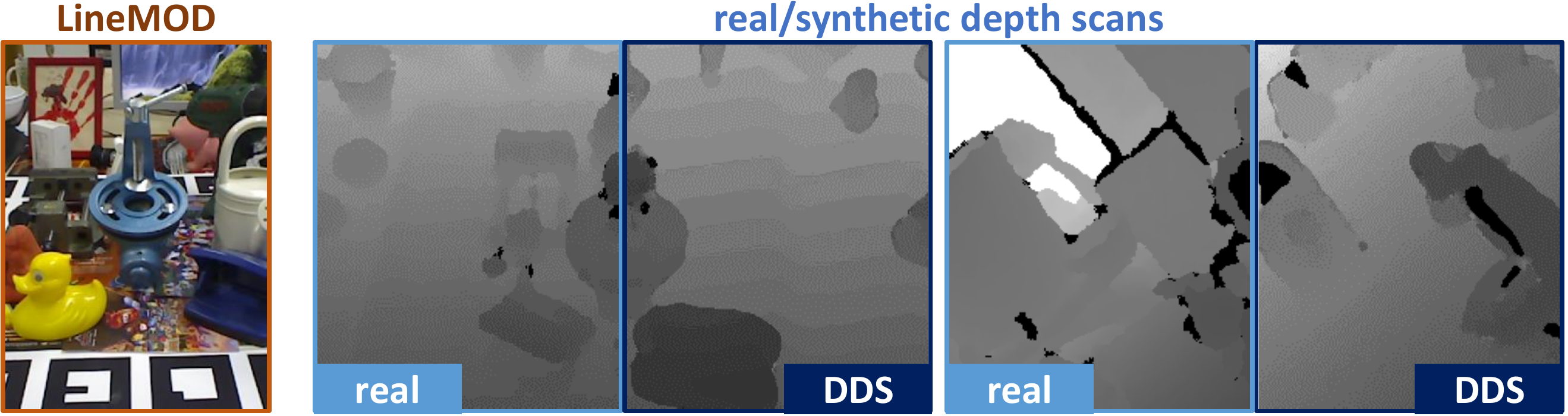}
\vspace{-1.6em}
\caption{\textbf{Differentiable Depth Sensor Simulation (\textit{DDS})  for the generation of highly-realistic depth scans.} \textit{DDS} works off-the-shelf, but can be further optimized unsupervisedly against real data, yielding synthetic depth scans valuable to the training of recognition algorithms (demonstrated here on LineMOD dataset~\cite{hinterstoisser2012model}).}
\label{fig:qual_linemod}  
\vspace{-1em}
\end{figure}

\begin{figure*}[ht!]
\centering
\includegraphics[width=\linewidth]{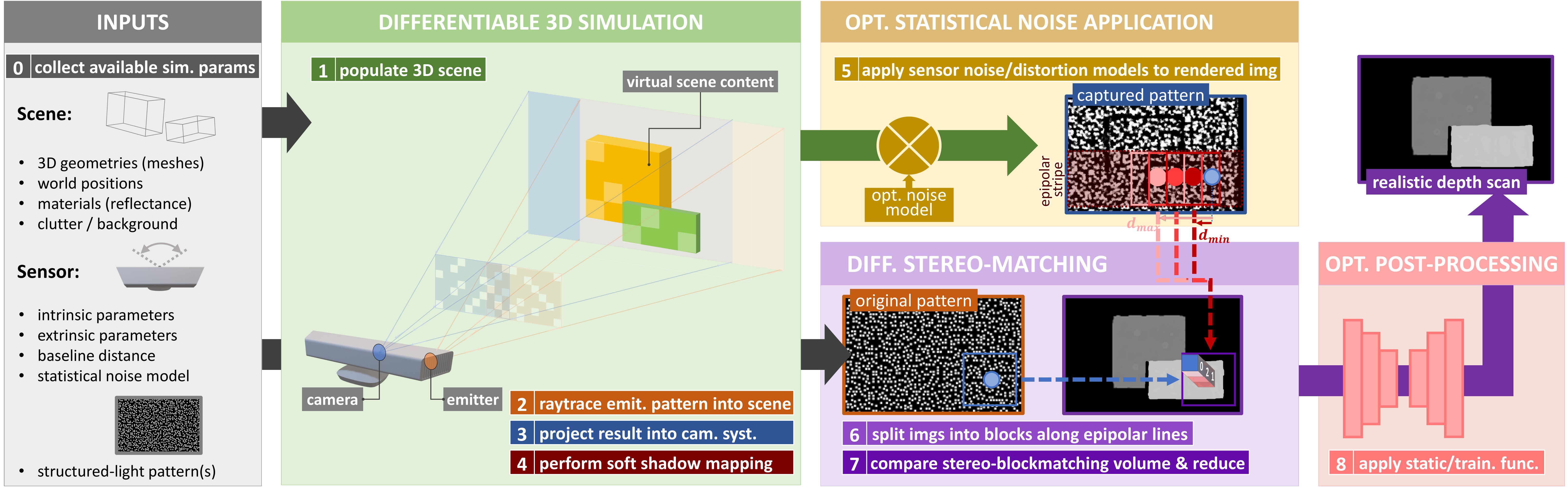}
\vspace{-1.6em}
\caption{\textbf{Pipeline overview.} \textit{DDS} differentiably simulates the physics and algorithmic mechanisms of real depth sensors.}
\label{fig:pipeline} 
\vspace{-.8em} 
\end{figure*}

The same cannot be said about depth-based applications. Unlike color camera that captures light intensity, structured-light depth sensors rely on stereo-vision mechanisms to measure the per-pixel distance between their focal plane and elements in the scene. They are useful for geometry-sensitive applications (\eg, robotics), but little effort has been made towards closing the realism gap \wrt synthetic depth (2.5D) scans or understanding their impact on the training of depth-based recognition methods. 
Some simulation pipelines~\cite{gschwandtner2011blensor,landau2015simulating,planche2017depthsynth} and domain adaptation schemes~\cite{tzeng2014deep,ganin2015unsupervised,tzeng2017adversarial,bousmalis2017unsupervised,zakharov2018keep,zakharov2019deceptionnet} have been proposed; but the former methods require extensive domain knowledge~\cite{planche2017depthsynth,zakharov2018keep} to be set up whereas some of the latter need relevant real images for their training~\cite{tzeng2014deep,ganin2015unsupervised,tzeng2017adversarial,bousmalis2016unsupervised}, and all fail to generalize to new sensors~\cite{gschwandtner2011blensor,landau2015simulating} or scenes~\cite{bousmalis2016unsupervised,zakharov2018keep}.

Borrowing from both simulation and learning-based principles, we propose herein a novel pipeline that virtually replicates depth sensors and can be optimized for new use-cases either manually (\eg, providing known intrinsic parameters of a new sensor) or automatically via supervised or unsupervised gradient descent (\eg, optimizing the pipeline over a target noise model or real scans). Adapting recent differentiable ray-tracing techniques~\cite{li2018differentiable,zhao2020physics,kato2020differentiable} and implementing novel \textit{soft} stereo-matching solutions, our simulation is differentiable end-to-end and can therefore be optimized via gradient descent, or integrated into more complex applications interleaving 3D graphics and neural networks. As demonstrated throughout the paper, our solution can off-the-shelf render synthetic scans as realistic as non-differentiable simulation tools~\cite{gschwandtner2011blensor,landau2015simulating,planche2017depthsynth}, outperforming them after unsupervised optimization. Applied to the training of deep-learning solutions for various visual tasks, it also outperforms unconstrained domain adaptation and randomization methods~\cite{tobin2017domain,bousmalis2017unsupervised,zakharov2018keep,zakharov2019deceptionnet}, \ie, resulting in higher task accuracy over real data; with a much smaller set of parameters to optimize.
In summary, our contributions are:

\noindent
\textbf{Differentiable Depth Sensor Simulation (\textit{DDS})} -- we introduce \textit{DDS}, an end-to-end differentiable, physics-based, simulation pipeline for depth sensors. As detailed in Section~\ref{sec:met}, \textit{DDS} reproduces the structured-light sensing and stereo-matching mechanisms of real sensors, off-the-shelf generating realistic 2.5D scans from virtual 3D scenes.

\noindent
\textbf{Optimizable Simulation through Gradient Descent} -- Because \textit{DDS} is differentiable \wrt most of the sensor and scene parameters, it can learn to better simulate new devices or approximate unaccounted-for scene properties in supervised or unsupervised settings. It can also be tightly incorporated within larger deep-learning pipeline, \eg, as a differentiable 3D-to-2.5D mapping function.

\noindent
\textbf{Benefits to Deep-Learning Recognition Methods} -- we demonstrate in Section~\ref{sec:exp} that \textit{DDS} is especially beneficial to recognition solutions that must rely on synthetic data. The various methods (for depth-based object classification, pose estimation, or segmentation) trained with \textit{DDS} performed significantly better when tested on real data, compared to the same methods trained with previous simulation tools or domain adaptation algorithms surveyed in Section~\ref{sec:rel}.

\section{Related work}
\label{sec:rel}

\paragraph{Physics-based Simulation for Computer Vision.}
Researchers have already demonstrated the benefits of physics-based rendering of color images to deep-learning methods~\cite{hodavn2019photorealistic,denninger2019blenderproc}, leveraging the extensive progress of computer graphics in the past decades. 
However, unlike color cameras, the simulation of depth sensors have not attracted as much attention. While it is straightforward to render synthetic 2.5D maps from 3D scenes (\cf \textit{z-buffer} graphics methods~\cite{strasser1974schnelle}), such \textit{perfect} scans do not reflect the structural noise and measurement errors impairing real scans, leaving recognition methods trained on this synthetic modality ill-prepared to handle real data~\cite{planche2017depthsynth,zakharov2018keep,planche2020bridging}. 

Early works~\cite{keller2009real,fallon2012efficient} tackling this \textit{realism gap} tried to approximate the sensors' noise with statistical functions that could not model all defects. 
More recent pipelines~\cite{gschwandtner2011blensor,landau2015simulating,planche2017depthsynth,reitmann2021blainder} are leveraging physics-based rendering to mimic the capture mechanisms of these sensors and render realistic depth scans, comprehensively modeling vital factors such as sensor noise, material reflectance, surface geometry, \etc. These works also highlighted the value of proper 2.5D simulation for the training of more robust recognition methods~\cite{planche2017depthsynth,planche2020bridging}.
However, extensive domain knowledge (\wrt sensor and scene parameters) is required to properly configured these simulation tools. Unspecified information and unaccounted-for phenomena (\eg, unknown or patented software run by the target sensors) can only be manually approximated, impacting the scalability to new use-cases.

With \textit{DDS}, we mitigate this problem by enabling the pipeline to learn missing parameters or optimize provided ones by itself.
This is made possible by the recent progress in differentiable rendering, with techniques modelling complex ray-tracing and light transport phenomena with continuous functions and adequate sampling~~\cite{loper2014opendr,li2018differentiable,zhao2020physics,kato2020differentiable}. More specifically, we build upon Li~\etal rendering framework~\cite{li2018differentiable} based on ray-tracing and Monte-Carlo sampling.

\begin{figure*}[t]
\centering
\includegraphics[width=\linewidth]{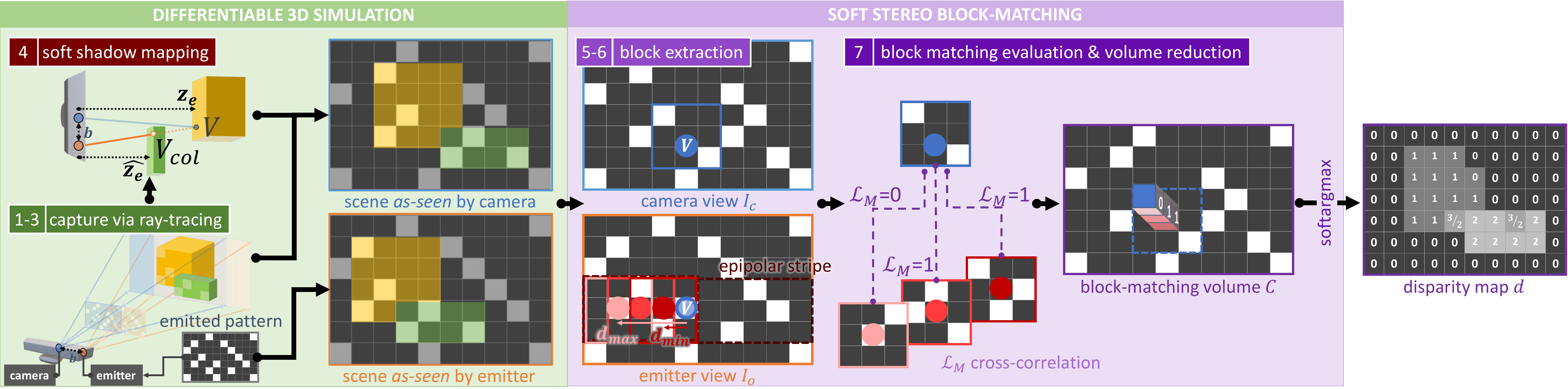}
\vspace{-1.6em}
\caption{\textbf{Gradient-based light transport and block-matching}, proposed in this paper to approximate the original methods.}
\label{fig:blockmatching} 
\vspace{-.8em} 
\end{figure*}

\paragraph{Domain Adaptation and Randomization.}
Similar to efforts \wrt color-image domains, scientists have also been proposing domain-adaptation solutions specific to depth data, replacing or complementing simulation tools to train recognition methods. 
Most solutions rely on unsupervised conditional generative adversarial networks (GANs)~\cite{goodfellow2014generative} to learn a mapping from synthetic to real image distributions~\cite{bousmalis2017unsupervised,yu2019simgan,lee2020drit++} or to extract features supposedly domain-invariant~\cite{ganin2016domain,zakharov2018keep}. Based on deep neural architectures trained on an unlabeled subset of target real data, these methods perform well over the specific image distribution inferred from these samples, but do not generalize beyond (\ie, they fail to map synthetic images to the real domain if the input images differ too much \wrt training data).

Some attempts to develop more scalable domain adaptation methods, \ie, detached from a specific real image domain (and therefore to the need for real training data), led to \textit{domain randomization} techniques~\cite{tobin2017domain}. These methods apply randomized transformations (handcrafted~\cite{tobin2017domain,zakharov2017,zakharov2018keep} or learned~\cite{zakharov2019deceptionnet}) to augment the training data, \ie, performing as an adversarial noise source that the recognition methods are trained against. The empirically substantiated claim behind is that, with enough variability added to the training set, real data may afterwards appear just as another noisy variation to the models.
We can, however, conceptually understand the sub-optimal nature of these unconstrained domain adaptation techniques, which consider any image transform in the hope that they will be valuable to the task, regardless of their occurence probability in real data.

By constraining the transforms and their trainable parameters to the optical and algorithmic phenomena actually impacting real devices, \textit{DDS} can converge much faster towards the generation of images that are both valuable to learning frameworks and photorealistic.

% Instead of simulation tools – or in complement of them – recent CNN-based methods are trying to further bridge the realism gap by learning a mapping from rendered to real data, directly in the image domain. Mostly based on unsupervised conditional generative adversarial networks (GANs) \rvs{[16], [17]} or style-transfer solutions \rvs{[18]}, these methods still need a set of real samples to learn their mapping. 
% %Adopting a GAN architecture too, we iterated over DepthSynth, adding a deep-learning-based post-processing step to the simulation pipeline. 
% Given real scans from the target depth sensor, a convolutional neural network is trained to edit the synthetic images to make them look more similar to their real equivalents. However, like earlier simulation methods, this post-processing step can only learn and apply statistical noise and failed to really bridge the realism gap. More precisely, by operating only on projected 2D images, the neural network does not have the capability to learn noise distributions linked to the properties of the original 3D scene (\eg, distance, orientation, and reflectance of the scene surfaces).
% Recent methods have focused on learning useful data augmentations which help improve the performance on the specific task the network is trained with. These methods have a CNN backbone and require to be trained along with the task adding training time and memory. 
% \rvs{DeceptionNet and DRIT/DRIT++ to be added}
\section{Methodology}
\label{sec:met}
% Similar to DepthSynth \cite{planche2017depthsynth}, the proposed pipeline simulates the different mechanisms of real depth sensors (c.f. \rvs{point to fig.}) to render realistic depth maps. Unlike DepthSynth, our novel method relies on differentiable operations at each step (pattern projection and capture, noise application, and stereo-matching). Like an artificial neural network, the pipeline is differentiable end-to-end and can, therefore, be fine-tuned/optimized over a training dataset to better achieve its task (the generation of realistic depth images).

% \rvs{figure caption. Figure to be added } Figure 18. Proposed solution for the differentiable simulation of structured-light depth sensors.

As illustrated in Figure~\ref{fig:blockmatching}, structured-light devices measure the scene depth in their field of view by projecting a light pattern onto the scene with their emitter. Their camera---tuned to the emitted wavelength(s)---captures the pattern's reflection from the scene.
Using the original pattern image $I_o$ and the captured one $I_c$ (usually filtered and undistorted) as a stereo signal, the devices infer the depth at every pixel by computing the discrepancy map between the images, \ie, the pixel displacements along the epipolar lines from one image to the other. 
The perceived depth $z$ can be directly computed from the pixel disparity $d$ via the formula $z = \frac{f_\lambda b}{d}$, with $b$ baseline distance between the two focal centers and $f_\lambda$ focal length shared by the device's emitter and camera. Note that depth sensors use light patterns that facilitate the discrepancy estimation, usually performed by block-matching algorithms~\cite{einecke2015multi,konolige1998small}.
Finally, most depth sensors perform some post-processing to computationally refine their measurements (\eg, using hole-filling techniques to compensate for missing data).

In this paper, we consider the simulation of structured-light depth sensors as a function $Z = G(\Phi)$, with $\Phi = \{\Phi_s, \Phi_c, \Phi_e\}$ set of simulation parameters. $G$ virtually reproduces the aforementioned sensing mechanisms, taking as inputs a virtual 3D scene defined by $\Phi_s$ (\eg, scene geometry and materials), the camera's parameters $\Phi_c$ (\eg, intrinsic and extrinsic values) and the emitter's $\Phi_e$ (\eg, light pattern image or function $\gamma_e$, distance $b$ to the camera); and returns a synthetic depth scan $Z$ \textit{as seen} by the sensor, with realistic image quality/noise.
We propose a simulation function $G$ differentiable \wrt $\Phi$, so that given any loss function $\mathcal{L}$ computed over $Z$ (\eg, distance between $Z$ and equivalent scan $\widehat{Z}$ from a real sensor), the simulation parameters $\Phi$ can be optimized accordingly through gradient descent.
The following section describes the proposed differentiable pipeline step by step, as shown in Figures~\ref{fig:pipeline} and \ref{fig:blockmatching}. 

\subsection{Pattern Capture via Differentiable Ray-Tracing}
To simulate realistic pattern projection and capture in a virtual 3D scene, we leverage recent developments in physics-based differentiable rendering~\cite{loper2014opendr,li2018differentiable,zhao2020physics,kato2020differentiable}.
Each pixel color $\gamma_c$ observed by the device camera is formalized as an integration over all light paths from the scene passing through the camera's pixel filter (modelled as a continuous function $k$), following the rendering equation:
\begin{equation}\label{eq:1}
    \gamma_c(\Phi) = \iiint k(x,y,\omega, \Phi_c)L(x,y,\omega\,;\,\Phi)\,dx\,dy\,d\omega,
\end{equation}
with $(x,y)$ continuous 2D coordinates in the viewport system, $\omega$ light path direction, and $L$ the radiance function modelling the light rays coming from the virtual scene (\eg, from ambient light and emissive/reflective surfaces)~\cite{li2018differentiable}. At any unit surface $V$ projected onto $(x,y)$ (in viewport coordinate system), the radiance $L$ with direction $\omega$ is, therefore, itself integrated over the scene content:
\begin{align}
    L(x,y,\omega\,;\,\Phi) = & \int_{\mathbb{S}^2} L_i(x,y,\omega\,;\,\Phi)f_s(V,\omega, \omega_i) \,d\sigma(\omega_i)
    \nonumber\\
    & + L_V(x,y,\omega\,;\,\Phi_s), \quad 
\end{align}
with $L_V$ radiance emitted by the surface (\eg, for the structured-light emitter or other light sources embodied in the scene), $L_i$ incident radiance, $f_s$ bidirectional reflectance distribution function~\cite{nicodemus1965directional}, $d\sigma$ solid-angle measure, and $\mathbb{S}^2$ unit sphere~\cite{zhao2020physics}.
As proposed by Li~\etal~\cite{li2018differentiable}, Monte Carlo sampling is used to estimate these integrals and their gradients: for continuous components of the integrand (\eg, inner surface shading), usual area sampling with automatic differentiation is applied, whereas discontinuities (\eg, surface edges) are handled via custom edge sampling.

More specific to our application, we simulate the structured-light pattern projection onto the scene and its primary contribution $L_e$ to $L$ for each unit surface $V$ as:
\begin{equation}
     L_e(x,y,\omega,\Phi) = \gamma_e(x_e, y_e,\Phi_e)\eta(V,\Phi_e), \quad 
\end{equation}
with $(x_e, y_e, z_e)^\top = M_e V$ projection of $V$ into the pattern image coordinate system defined by the projection matrix $M_e$,  $\gamma_e$ continuous representation of the structured-light pattern emitted by the sensor, and $\eta$ light intensity (\eg, as a function of the distance to the emitter). In other words, for surfaces visible to the camera, we trace rays from them to the light emitter to measure which elements of its pattern are lighting the surfaces (\cf steps 1-3 in Figure~\ref{fig:blockmatching}).

As highlighted in various studies~\cite{landau2015simulating,landau2016,planche2017depthsynth,planche2020bridging}, due to the baseline distance between their emitter and camera, depth sensors suffer from shadow-related capture failure, \ie, when a surface $V$ contributing to $\gamma_c$ does not receive direct light from the emitter due to occlusion of the light rays by other scene elements (\cf step 4 in Figure~\ref{fig:blockmatching}). Therefore, we propose a soft \textit{shadow mapping} procedure~\cite{williams1978casting,akenine2019real} that we model within the light intensity function $\eta$ as follows:
\begin{equation}
     \eta(V) = \frac{\eta_c}{z_e^2}\left(1 - \sigma(z_e - \widehat{z_e} - \xi)\right),
\end{equation}
with $\sigma$ sigmoid operator (replacing the discontinuous step function used in traditional shadow mapping), $\eta_c$ emitter intensity, and $\widehat{z_e}$ computed as $(x_e, y_e, \widehat{z_e})^\top = M_e V_{col}$ where $V_{col}$ is the first surface hit by the virtual ray thrown from the emitter focal center toward $V$ (\ie, $V_{col}$ superposed to $V$ but closer in the emitter 2D coordinate system). We add an optimizable bias $\xi \in \mathbb{R}$ to prevent \textit{shadow acne} (shadow artifacts due to distance approximations)~\cite{LearnOpenGLShadowMapping}.

Estimating $\gamma_c(\Phi)$ accounting for the scene and sensor properties $\Phi$, we obtain the rasterized image $I_c$. To cover non-modelled physics phenomena (\eg, lens defects) and according to previous works~\cite{gschwandtner2011blensor,planche2017depthsynth}, we also adopt an optional noise function $f_n$ applied to $I_c$, \eg, $f_n(I_c) = I_c + \Delta I$, with $\Delta I = \epsilon\sigma_n + \mu_n$, $\{\mu_n, \sigma_n\} \in \Phi_c$, and $\epsilon \sim \mathcal{N}(0, 1)$ (\cf reparameterization trick~\cite{fabius2015variational,makhzani2015adversarial}).

% \begin{figure*}[t]
% \centering
% \includegraphics[width=\linewidth]{figures/figure_qualitative_results_2d3ds.pdf}
% \vspace{-1.6em}
% \caption{\textbf{Qualitative comparison of simulated scans.} Synthetic depth images rendered from reconstructed 3D indoor scenes of the \textit{2D-3D-Semantic} dataset~\cite{armeni2017joint}, compared to real scans from the Matterport Pro2 sensor. Note that the Pro2 device relies on 3 stacked depth sensors, hence the high accuracy and reduced shadow noise.}
% \label{fig:qual_2d3ds}  
% \vspace{-.8em} 
% \end{figure*}

\begin{figure*}[t]
\centering
\includegraphics[width=0.95\linewidth]{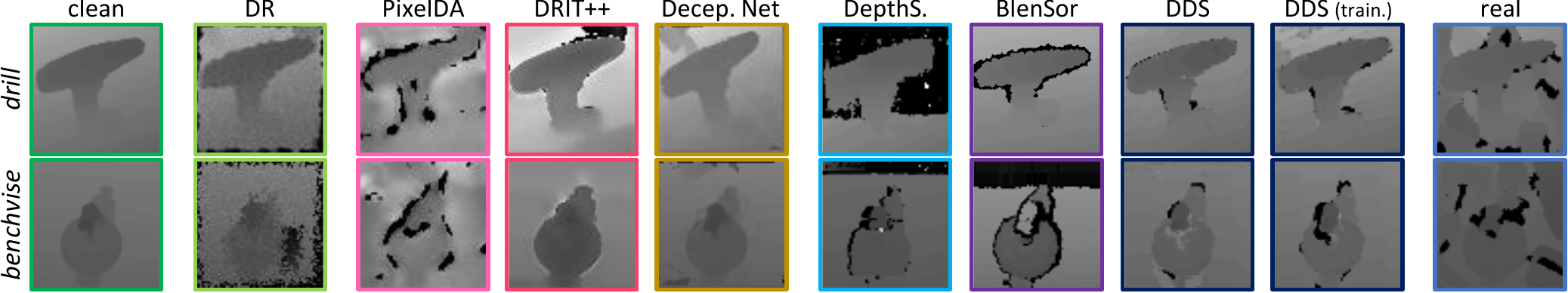}
\vspace{-.6em}
\caption{\textbf{Domain adaptation and simulation results}, on \textit{Cropped LineMOD}~\cite{hinterstoisser2012model,bousmalis2017unsupervised,zakharov2019deceptionnet} (real scene clutter not reproduced).}
\label{fig:linemod_comp} 
\vspace{-.8em} 
\end{figure*}

\subsubsection{Differentiable Stereo Block-Matching}
Similar to real depth sensors, our pipeline then compares the computed $I_c$ with a rasterized version $I_o$ of the original pattern (both of size $H \times W$) to identify stereo-correspondences and infer the disparity map.
Differentiable solutions to regress disparity maps from stereo signals have already been proposed, but these methods rely on CNN components to perform their task either more accurately~\cite{luo2016efficient,chang2018pyramid,duggal2019deeppruner} or more efficiently~\cite{kendall2017end}. Therefore, they are bound to the image domain that they were trained over. 
%Increasingly accurate and computationally efficient, these methods are however learning-based (built upon deep-learning algorithms) and can perform their task properly only over the specific image domain that they were trained for. 
Since our goal is to define a scene-agnostic simulation pipeline, we proposed instead an improved  continuous implementation~\cite{kendall2017end} of the classic stereo block-matching algorithm applied to disparity regression~\cite{konolige1998small,konolige2010projected}, illustrated in Figure~\ref{fig:blockmatching}.
The algorithm computes a matching cost volume $C \in \mathbb{R}^{H \times W \times N_d}$ by sliding a $w \times w$ window over the two images, comparing each block in $I_c$ with the set of $N_d$ blocks in $I_o$ extracted along the same epipolar line. 
Considering standard depth sensors with the camera and emitter's focal planes parallel, the epipolar lines appear horizontal in their image coordinate systems (with $N_d = W$), simplifying the equation into:
\begin{equation}
    C(x, y, \delta) = \sum_{i=x+u}^{x+w}\sum_{j=y+v}^{y+w} \mathcal{L}_M\left(I_{c;\,i,j}, I_{o;\,i,j-\delta} \right),
\end{equation}
with $\delta \in [y-N_d-w, y]$ horizontal pixel displacement and $\mathcal{L}_M$ matching function (we opt for cross-correlation). Matrix unfolding operations are applied to facilitate volume inference.
Formulating the task as a \textit{soft} correspondence search, we reduce $C$ into the disparity map $d$ as follows:
%\begin{equation}
$    d(x, y) = \softargmax_\delta C(x, y, \delta) $
%\end{equation}
with $\softargmax_i X = \sum_i \frac{ie^{\beta X_i}}{\sum_i e^{\beta X_i}}$
and $\beta \in \mathbb{R}$ optimizable parameter controlling the temperature of the underlying probability map.
From this, we can infer the simulated depth scan $Z = \frac{f_\lambda b}{d}$.

However, as it is, the block-matching method would rely on an excessively large cost volume $H \times W \times W$ (\ie, with $N_d = W$) making inference and gradient computation impractical. We optimize the solution by considering the measurement range $[z_{min}, z_{max}]$ of the actual sensor (\eg, provided by the manufacturer or inferred from focal length), reducing the correspondence search space accordingly, \ie, with $\delta \in [d_{min}, d_{max}] = [\round{\frac{f_\lambda b}{z_{max}}}, \round{\frac{f_\lambda b}{z_{min}}}]$ (dividing $N_d$ tenfold for most sensors). The effective disparity range can be further reduced, \eg, by considering the min/max \textit{z-buffer} values in the target 3D scene.

The computational budget saved through this scheme can instead be spent refining the depth map. Modern stereo block-matching algorithms perform fine-tuning steps to achieve sub-pixel disparity accuracy, though usually based on global optimization operations that are not directly differentiable~\cite{humenberger2010fast,michael2013real}.
To improve the accuracy of our method without trading off its differentiability, we propose the following method adapted from~\cite{landau2015simulating}:
Let $n_{sub}$ be an hyperparameter representing the desired pixel fraction accuracy. We create $\left\{I_{o,i}\right\}_{i=1}^{n_{sub}}$ lookup table of pattern images with a horizontal shift of $i/n_{sub}$ px. Each $I_{o,i}$ is pre-rendered (once) via Equation~\ref{eq:1} with $\Phi_{s,i}$ defining a virtual scene containing a single flat surface parallel to the sensor focal planes placed at distance 
$\frac{f_\lambda b}{d_{min,i}}$ with $d_{min,i} = d_{min}+\frac{i}{n_{sub}}$ 
% $z_{max} + \frac{n_{sub}}{i}f_\lambda b$ 
(hence a global disparity of $i/n_{sub}$ between $I_o$ and $I_{o,i}$). At simulation time, block-matching is performed between $I_c$ and each $I_{o,i}$, interlacing the resulting cost volumes and reducing them at once into the refined disparity map.

Finally, similar to the noise function optionally applied to $I_c$ after capture, our pipeline allows $Z$ to be post-processed, if non-modelled functions need to be accounted for (\eg, device's hole-filling operation). In the following experiments, we present different simple post-processing examples (none, normal noise, or shallow CNN).
	
\section{Experiments}
\label{sec:exp}

Through various experiments, we evaluate the photorealism of depth images rendered by \textit{DDS} and their value \wrt training recognition method or solving inverse problems.

\subsection{Realism Study}\label{sec:exp-realismstudy}
First, we qualitatively and quantitatively compare \textit{DDS} results with real sensor scans and data from other pipelines.

% \begin{figure*}[t]
% \centering
% \includegraphics[width=\linewidth]{figures/figure_noisestudy2.pdf}
% \vspace{-1.7em}
% \caption{\textbf{Quantitative sensor noise study \wrt radial distance $r$.} Standard depth error as a function of $r$ distance to the focal center in the image system, plotted for actual \textit{Kinect V1} scans and simulated depth images of a flat surface placed at various distances $z$ from the real or virtual sensor. Scans from \textit{DDS} show the same noise trends as the real sensor.}
% \label{fig:noise_study_raddist}  
% \vspace{-.8em} 
% \end{figure*}
\begin{figure}[t]
\centering
\includegraphics[width=\linewidth]{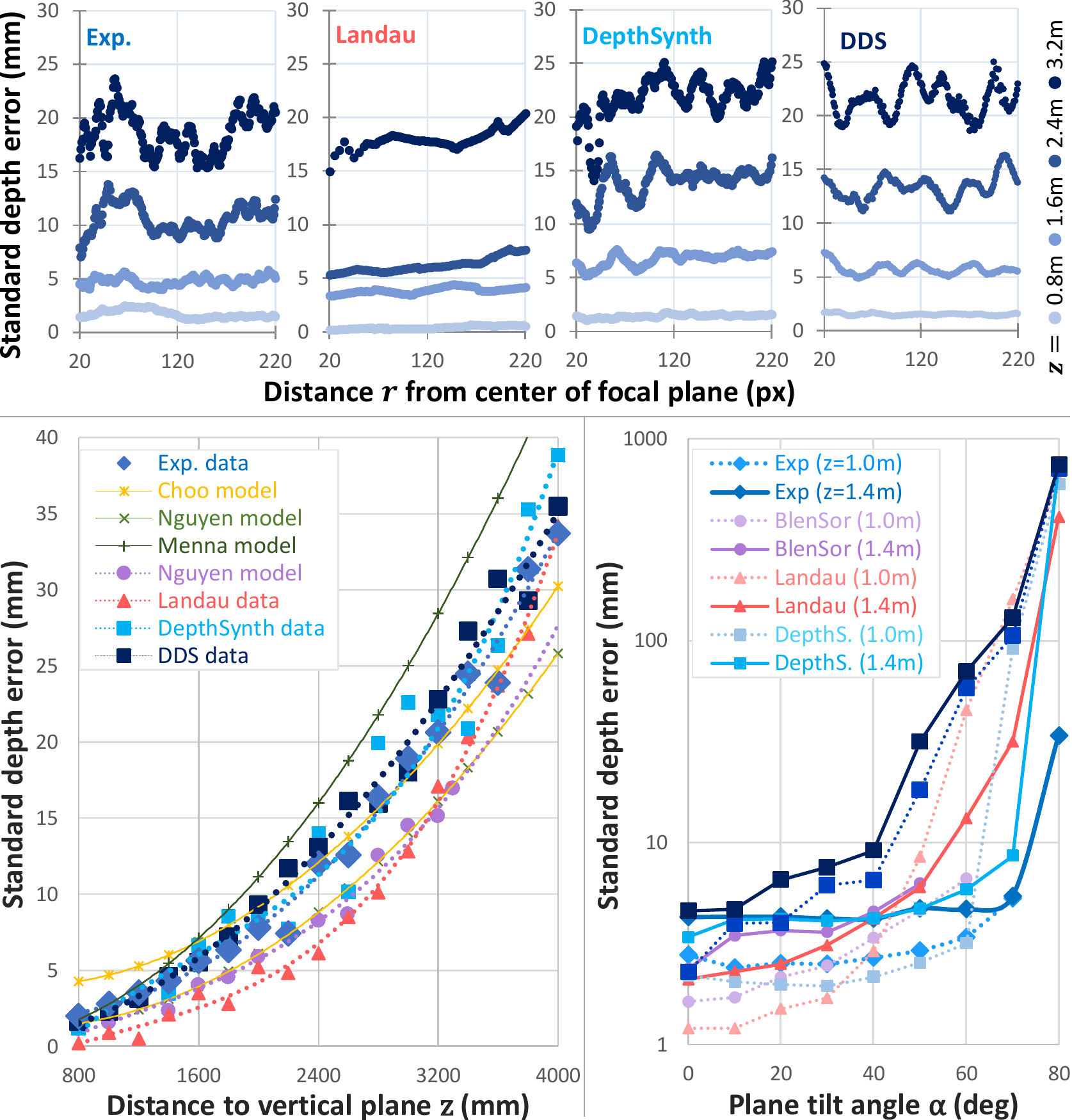}
\vspace{-1.7em}
\caption{\textbf{Sensor noise study.} 
Given a flat surface placed at various distances $z$ and tilt angles $\alpha$ \wrt the sensor, we plot the standard depth error as a function of $r$ distance to the focal center in screen space, of $z$, and of $\alpha$; for actual and simulated \textit{Kinect V1} scans and statistical sensor models.}
\label{fig:noise_study_raddist}  
\vspace{-.8em} 
\end{figure}

% \begin{figure}[t]
% \centering
% \includegraphics[width=\linewidth]{figures/figure_noisestudy.pdf}
% \vspace{-1.7em}
% \caption{\textbf{Sensor noise study \wrt depth and surface angle.} Standard depth error of real and simulated \textit{Kinect} images or statistical sensor models for a flat surface placed at various distances $z$ and tilt angle $\alpha$ \wrt the sensor.}
% \label{fig:noise_study_distandangle}  
% \vspace{-.8em} 
% \end{figure}

\paragraph{Qualitative Comparison.}
Visual results are shared in Figures~\ref{fig:qual_linemod} and \ref{fig:linemod_comp} (\wrt \textit{Microsoft Kinect V1} simulation), 
%and Figure~\ref{fig:qual_2d3ds} (\wrt \textit{Matterport Pro2}), 
as well as in the supplementary material (\wrt \textit{Matterport Pro2}). 
We can observe that off-the-shelf \textit{DDS} reproduces the image quality of standard depth sensors (\eg, \textit{Kinect V1}): \textit{DDS} scans contain shadow noise, quantization noise, stereo block-mismatching, \etc, similar to real images and previous simulations~\cite{gschwandtner2011blensor, planche2017depthsynth} (\cf empirical study of depth sensors' noise performed by Planche~\etal~\cite{planche2017depthsynth}).
Figure \ref{fig:linemod_comp} and supplementary material further highlight how, unlike static simulations, ours can learn to tune up or down its inherent noise to better model sensors of various quality.
% Figure~\ref{fig:qual_2d3ds} further highlights how, unlike static simulations, the proposed solution can learn to tune down its inherent noise to model more precise sensors such as the multi-shot \textit{Matterport} device (composed of 3 sensors).

\paragraph{Quantitative Comparison.}
Reproducing the experimental protocol of previous 2.5D simulation methods~\cite{landau2016, planche2017depthsynth}, we statistically model the depth error incurred by \textit{DDS} as function of various scene parameters, and compare with empirical and statistical models from real sensor data. 

\noindent
\textbullet~\textit{Protocol.} 
Studying the \textit{Microsoft Kinect V1} sensor, Landau~\etal~\cite{landau2015simulating,landau2016} proposed the following protocol (further illustrated in the supplementary material). In real and simulated world, a flat surface is placed in front of the sensor. The surface is considered as a plane $(P, \overrightarrow{u}, \overrightarrow{v})$  with $P=(0, 0, z)$, $\overrightarrow{u} = (1, 0, 0)$, and $\overrightarrow{v} =( 0, \sin{\alpha}, \cos{\alpha})$ in camera coordinate system (\ie, a plane at distance $z$ and tilt angle $\alpha$ \wrt focal plane).
For each image captured in this setup, the standard depth error for each pixel $q$ is computed as function of the distance $z$, the tilt angle $\alpha$, and the radial distance $r$ to the focal center. Like Landau~\etal~\cite{landau2015simulating,landau2016} and Planche~\etal~\cite{planche2017depthsynth}, we compare the noise functions of our method with those of the actual \textit{Kinect V1} sensor, as well as the noise functions computed for other state-of-the-art simulation tools (\textit{BlenSor}~\cite{gschwandtner2011blensor}, Landau's~\cite{landau2015simulating}, and \textit{DepthSynth}~\cite{planche2017depthsynth}) and noise models proposed by researchers studying this sensor (Menna~\etal~\cite{menna2011geometric}, Nguyen~\etal~\cite{nguyen2012modeling} and Choo~\etal~\cite{choo2014statistical,landau2016}).
	
\noindent
\textbullet~\textit{Results.} 
In Figure~\ref{fig:noise_study_raddist}, we first plot the error as a function of the radial distance $r$ to the focal center. \textit{DDS} performs realistically: like other physics-based simulations~\cite{gschwandtner2011blensor,planche2017depthsynth}, it reproduces the noise oscillations, with their amplitude increasing along with distance $z$---a phenomenon impairing real sensors, caused by pattern distortion.

We also plot the standard error as a function of the distance $z$ and of the incidence angle $\alpha$. While our simulated results are close to the real ones \wrt distance, we can observe that noise is slightly over-induced \wrt tilt angle. The larger the angle, the more stretched the pattern appears on the surface, impairing the block-matching procedure. Most algorithms fail matching overly-stretched patterns (\cf exponential error in the figure), but our custom differentiable block-matching solution is unsurprisingly less robust to block skewing than the multi-pass methods used in other simulations~\cite{gschwandtner2011blensor,planche2017depthsynth}. This could be tackled by adopting some more advanced block-matching strategies from the literature and rewriting them as continuous functions. This would however increase the computational footprint of the overall simulation and would only benefit applications where high photorealism is the end target. In the next experiments, we instead focus on deep-learning applications.

\subsection{Applications to Deep Learning}
We now illustrate how deep-learning solutions can benefit from our simulation method. We opt for various key recognition tasks over standard datasets, comparing the performance of well-known CNNs as a function of the data and the domain adaptation framework used to train them.

\paragraph{2.5D Semantic Segmentation.}

We start by comparing the impact of simulation tools on the training of a standard CNN for depth-based semantic segmentation.

\noindent
\textbullet~\textit{Dataset.} For this task, we choose the \textit{2D-3D-Semantic} dataset by Armeni~\etal~\cite{armeni2017joint} as it contains RGB-D indoor scans shot with a \textit{Matterport Pro2} sensor, as well as the camera pose annotations and the reconstructed 3D models of the 6 scenes. It is, therefore, possible to render synthetic images aligned with the real ones. We split the data into training/testing sets as suggested by \textit{2D-3D-S} authors~\cite{armeni2017joint} (fold \#1, \ie, 5 training scenes and 1 testing one). For the training set, we assume that only the 3D models, images and their pose labels are available (not the ground-truth semantic masks). Note also that for the task, we consider only the 8 semantic classes (out of 13) that are discernible in depth scans (\eg, \textit{board} are indistinguishable from \textit{wall} in 2.5D scans) and present in the training scenes.

\begin{table}[t]
\centering
\caption{\textbf{Comparative study \wrt training usage}, measuring the accuracy of a CNN~\cite{he2016deep,wang2018understanding,wu2019wider} performing semantic segmentation on real 2.5D scans from the indoor \textit{2D-3D-S} dataset~\cite{armeni2017joint}, as a function of the method used to render its training data ($\uparrow$ = the higher the value, the better).
}
\label{tab:2d3dds} 
\vspace{-.7em} 
\resizebox{1\linewidth}{!}{
\def\arraystretch{1}% \label{tab:gan_comp}
\begin{tabu}{@{}c|llllllll|c@{}}
\toprule
\multicolumn{1}{c}{\multirow{2}{*}{\shortstack{\\\textbf{Train.}\\\textbf{Data}\\\textbf{Source}}}} & 
\multicolumn{8}{|c}{\textbf{Mean Intersection-Over-Union (mIoU)$^\uparrow$}} & 
\multicolumn{1}{|c}{\multirow{2}{*}{\shortstack{\\[0.2em]\textbf{Pixel}\\\textbf{Acc.$^\uparrow$}}}}
\\
\cmidrule(lr){2-9}
& 

\parbox[t]{1mm}{\rotatebox[origin=c]{45}{\footnotesize\textbf{\textit{bookc.}}}}&
\parbox[t]{2mm}{\rotatebox[origin=c]{45}{\footnotesize\textbf{\textit{ceili.}}}}&
\parbox[t]{2mm}{\rotatebox[origin=c]{45}{\footnotesize\textbf{\textit{chair}}}}&
\parbox[t]{2mm}{\rotatebox[origin=c]{45}{\footnotesize\textbf{\textit{clutter}}}}&
\parbox[t]{2mm}{\rotatebox[origin=c]{45}{\footnotesize\textbf{\textit{door}}}}&
\parbox[t]{2mm}{\rotatebox[origin=c]{45}{\footnotesize\textbf{\textit{floor}}}}&
\parbox[t]{2mm}{\rotatebox[origin=c]{45}{\footnotesize\textbf{\textit{table}}}}&
\parbox[t]{2mm}{\rotatebox[origin=c]{40}{\footnotesize\textbf{\textit{wall}}}}
% & \textbf{mean}  

& \\

\midrule

clean 
& .003 & .018 & .002 & .087 & .012 & .052 & .091 & .351
% & .085
& 35.3\%

\\\cmidrule(lr){1-10}
\textit{BlenSor}{\scriptsize~\cite{gschwandtner2011blensor}} 
& .110 & .534 & .119 & .167 & .148 & .561 & .082 & .412
% & .161
& 51.6\%

\\
\textit{DepthS.}{\scriptsize~\cite{planche2017depthsynth}}
& .184 & .691 & .185 & .221 & .243 & .722 & .235 & .561
% & .264
& 65.3\%

\\
\textit{DDS} 
& .218 & .705 & .201 & .225 & .240 & .742 & .259 & .583
% & .278
& 62.9\%

\\
\textit{DDS} \footnotesize{(train.)} 
& \textbf{.243} & .711 & \textbf{.264} & .255 & .269 & .794 & .271 & .602
% & .296
& 69.8\%

\\\cmidrule(lr){1-10}
% \textit{DDS} + real 
% & XXX & XXX & XXX & XXX & XXX & XXX & XXX & XXX
% % & XXX
% & XXX\%

% \\
real 
& .135 & \textbf{.770} & .214 & \textbf{.277} & \textbf{.302} & \textbf{.803} & \textbf{.275} & \textbf{.661}
% & \textbf{.301}
& \textbf{73.5\%}

\\
\bottomrule
\end{tabu}
\vspace{-1em} 
}
%	\vspace{0.4cm}   
\vspace{-1em} 
\end{table}

\noindent
\textbullet~\textit{Protocol.} Using the 3D models of the 5 training scenes, we render synthetic 2.5D images and their corresponding semantic masks using a variety of methods from the literature~\cite{aldoma2012point,gschwandtner2011blensor,planche2017depthsynth}. \textit{DDS} is both applied off-the-shelf (only entering the \textit{Pro2} sensor's intrinsic information), and after being optimized via supervised gradient descent (combining Huber and depth-gradient losses~\cite{huber1992robust,jiao2018look}) against the real scans from one training scene (scene \#3).  Each synthetic dataset, and the dataset of real scans as upper-bound target, is then used to train an instance of a standard ResNet-based CNN~\cite{he2016deep,wang2018understanding,wu2019wider} for semantic segmentation (we choose the \textit{Dice} loss to make up for class imbalance~\cite{drozdzal2016importance}).

\noindent
\textbullet~\textit{Results.} We measure the performance of each model instance in terms of per-class mean intersection-over-union~\cite{jaccard1912distribution,rahman2016optimizing} and pixel accuracy. Results are shared in Table~\ref{tab:2d3dds}.
We can observe how data from both untrained and trained \textit{DDS} result in the most accurate recognition models (among those trained on purely synthetic data), with values on par or above those of the models trained on real annotated data for some classes. Even though \textit{DDS} may not perfectly simulate the complex, multi-shot \textit{Matterport} sensor, its ability to render larger and more diverse datasets can be easily leveraged to achieve high recognition accuracy. 

\paragraph{Classification and Pose Estimation.}
We now perform an extensive comparison, as well as partial ablation study, \wrt the ubiquitous computer vision task of instance classification and pose estimation (ICPE)~\cite{Wohlhart15,bousmalis2017unsupervised,zakharov2017,zakharov2018keep}.

\noindent
\textbullet~\textit{Dataset.}
For this task, we select the commonly-used \textit{Cropped LineMOD} dataset~\cite{hinterstoisser2012model,Wohlhart15,bousmalis2017unsupervised}, composed of $64\times64$ RGB-D image patches of 11 objects under various poses, captured by a \textit{Kinect V1} sensor, in cluttered environments. Disregarding the RGB modality for this experiment, we split the dataset into a non-annotated training set $X^r_{trn}$ of 11,644 depth images, and a testing set $X^r_{tst}$ of 2,919 depth images with their class and pose labels. The \textit{LineMOD} dataset also provides a reconstructed 3D model of each object, used to render annotated synthetic training images. For fair comparison, all 3D rendering methods considered in this experiment are provided the same set of 47,268 viewpoints from which to render the images. These viewpoints are sampled from a virtual half-icosahedron centered on each target object, with 3 different in-plane rotations (\ie, rotating the camera around its optical axis)~\cite{Wohlhart15,zakharov2017,zakharov2018keep,planche2019seeing}.

\noindent
\textbullet~\textit{Protocol.}
For this experiment, we opt for the generic task CNN from~\cite{ganin2015unsupervised}, trained for object classification and rotation estimation via the loss 
$\mathcal{L}_{icpe} = \mathbb{E}_{x,(y,q)}\left[-y^\top\log{\hat{y}} + \xi \log{\left(1 - |q^\top \hat{q}|\right)}\right]$, where the first term is the class-related cross-entropy and the second term is the log of a 3D rotation metric for quaternions~\cite{bousmalis2017unsupervised,zakharov2019deceptionnet}, with $\xi$ pose loss factor, $x$ input depth image, $\{y, q\}$ resp. ground-truth one-hot class vector and quaternion, and $\{\hat{y}, \hat{q}\}$ resp. predicted values.
Again, we measure the network's classification accuracy and rotational error as a function of the data that it was trained on, extending the comparison to different online or offline augmentation and domain adaptation schemes (\cf Figure~\ref{fig:linemod_comp} for visual comparison).

\begin{figure*}[t]
\centering
\includegraphics[width=1\linewidth]{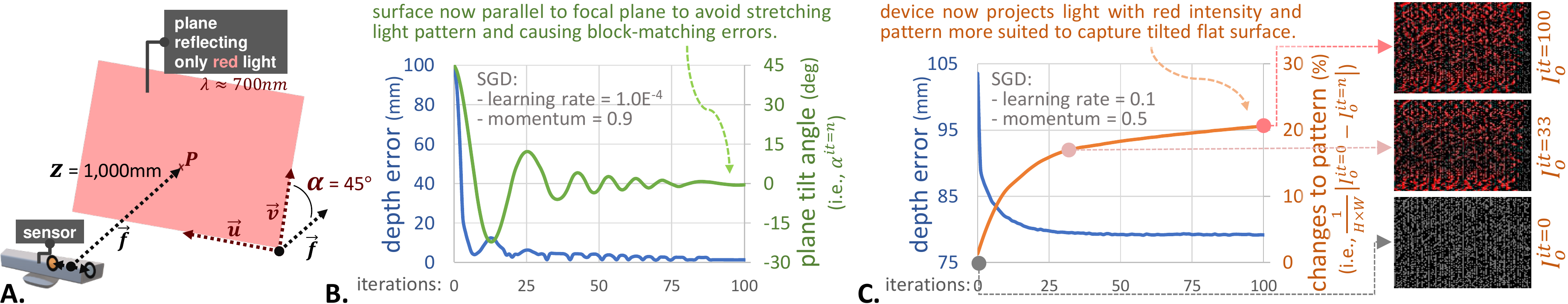}
\vspace{-1.7em}
\caption{\textbf{Optimization of scene and sensor parameters} via simulation, to improve sensor precision in controlled use-cases.
A.\ Experimental setup \cf Section~\ref{sec:exp-realismstudy}; B.\ Optimization of the scene parameters (\eg, pose) to reduce sensor standard error; C.\ Optimization of the sensor (\eg pattern structure and frequencies) to improve its accuracy \wrt such scenes/materials.}
\label{fig:scene_sensor_optim} 
\vspace{-.8em} 
\end{figure*}

For domain adaptation solutions such as \textit{PixelDA}~\cite{bousmalis2017unsupervised} and \textit{DeceptionNet}~\cite{zakharov2019deceptionnet}, the recognition network $T$ is trained against a generative network $G$ whose task is to augment the input synthetic images before passing them to $T$. This adversarial training framework, with $G$ trained unsupervisedly against $T$~\cite{zakharov2019deceptionnet} and/or a discriminator network $D$~\cite{bousmalis2017unsupervised,zakharov2018keep} using non-annotated real images $X^r_{trn}$, better prepares $T$ for its task on real data, \ie, training it on noisier and/or more realistic synthetic images. To further demonstrate the training of our simulation, this time in a less constrained, unsupervised setting, we reuse \textit{PixelDA} training framework, replacing its \textit{ResNet}-based~\cite{he2016deep} generator by \textit{DDS}. Our method is, therefore, unsupervisedly trained along with the task network, so that \textit{DDS} learns to render synthetic images increasingly optimized to help $T$ with its training. Three instance of \textit{DDS} are thus compared: (a) off-the-shelf, (b) with $\Phi = \{\xi, \mu_n, \sigma_n, \beta\}$ (\ie, parameters \wrt shadows, normal noise, and softargmax) optimized unsupervisedly, and (c) same as the previous but adding 2 trainable convolution layers as post-processing ($|\Phi|$ = 2,535 only in total).

\noindent
\textbullet~\textit{Results.} 
Table~\ref{tab:linemod} presents a detailed picture of state-of-the-art training solutions for scarce-data scenarios (basic or simulation-based image generation, static or GAN-based offline or online image transformations, \etc) and their performance on the task at hand. The various schemes are further sorted based on their requirements \wrt unlabeled real images and on the size of their parameter space.

\begin{table}[t]
\centering
\caption{\textbf{Comparative and ablative study}, measuring the impact of unsupervised domain adaptation, sensor simulation (Sim), and domain randomization (DR, \ie, using randomized 2.5D transforms \cf \cite{zakharov2018keep,zakharov2019deceptionnet}) on the training of a CNN~\cite{ganin2015unsupervised} for depth-based instance classification and pose estimation on the \textit{Cropped LineMOD} dataset~\cite{hinterstoisser2012model,bousmalis2017unsupervised,zakharov2019deceptionnet}.
}
\label{tab:linemod} 
\vspace{-.7em} 
\resizebox{1\linewidth}{!}{
\def\arraystretch{1}% \label{tab:gan_comp}
\begin{tabu}{@{}cc|cc|cc|cc@{}}
\toprule
\multicolumn{2}{c}{\multirow{2}{*}{}} & 
\multicolumn{2}{|c}{\textbf{Augmentations}} & 
\multicolumn{2}{|c}{\textbf{Sim/DA Req.}} &  
\multicolumn{1}{|c}{\multirow{2}{*}{\shortstack{\textbf{Class.}\\\textbf{Accur.}$^\uparrow$}}} &
\multicolumn{1}{c}{\multirow{2}{*}{\shortstack{\textbf{Rot.}\\\textbf{Error}$^\downarrow$}}}
\\
\cmidrule(lr){3-4}\cmidrule(lr){5-6}
& & \textbf{offline} & \textbf{online} & $\bm{X^r_{trn}}$ &$\bm{|\Phi|}$ &\\

\midrule

&\multicolumn{1}{c|}{\multirow{2}{*}{Basic}} &
        &	&	            &	      & 46.8\% &	67.0$^{\circ}$
\\
&&      & DR&               &	      & 70.7\% &	53.1$^{\circ}$
\\
\cmidrule(lr){1-8}

\parbox[t]{2mm}{\multirow{5}{*}{\rotatebox[origin=c]{90}{Dom. Adap.}}} &
\multirow{1}{*}{\textit{PixelDA}}{\footnotesize~\cite{bousmalis2017unsupervised}}   & 
        & GAN& $\checkmark$	& 1.96M &	85.7\% &	40.5$^{\circ}$	
\\
\cmidrule(lr){2-8}

& 
\multirow{2}{*}{\textit{DRIT++}{\footnotesize~\cite{lee2020drit++}}}
&   GAN &   & $\checkmark$  & 12.3M &   68.0\% &	60.8$^{\circ}$	\\
&&  GAN & DR & $\checkmark$  & 12.3M &   87.7\% &	39.8$^{\circ}$	\\
\cmidrule(lr){2-8}

&
\multirow{1}{*}{\textit{Decep.Net}{\footnotesize~\cite{zakharov2019deceptionnet}}} & 
        & DR & 				& 1.54M &	80.2\% &	54.1$^{\circ}$	
\\
\cmidrule(lr){1-8}
\parbox[t]{2mm}{\multirow{7}{*}{\rotatebox[origin=c]{90}{Sensor Sim.}}}  &
\multirow{2}{*}{\textit{DepthS.}{\footnotesize~\cite{planche2017depthsynth}}} &
   Sim     & 	&	        &	      & 71.5\% &	52.1$^{\circ}$	\\
&&   Sim     & DR	&	        &	      & 76.6\% &	45.4$^{\circ}$	\\
\cmidrule(lr){2-8}

&
\multirow{2}{*}{\textit{BlenSor}{\footnotesize~\cite{gschwandtner2011blensor}}} &
   Sim     & 	&	        &	      & 67.5\% &	63.4$^{\circ}$	\\
&&   Sim     & DR	&	        &	      & 82.6\% &	41.4$^{\circ}$	\\
\cmidrule(lr){2-8}

&
\multirow{2}{*}{\shortstack{\textit{DDS}\\(untrained)}} &
   Sim     & 	&	        &	      & 69.7\% &	67.6$^{\circ}$	\\
&&   Sim     & DR	&	        &	      & 89.6\% &	39.7$^{\circ}$	\\
\cmidrule(lr){1-8}

\parbox[t]{2mm}{\multirow{6}{*}{\rotatebox[origin=c]{90}{Combined}}} &
\multirow{4}{*}{\textit{DDS}} &
   Sim     & 	& $\checkmark$ & 4	  & 81.2\% &	49.1$^{\circ}$	\\
&& Sim     & DR	& $\checkmark$ & 4	  & 90.5\% &	39.4$^{\circ}$	\\
&& Sim\scriptsize{+conv} & 	& $\checkmark$ & 2,535	  & 85.5\% &	45.4$^{\circ}$	\\
&& Sim\scriptsize{+conv} & DR&$\checkmark$ & 2,535	  & 93.0\% &	31.3$^{\circ}$	\\
\cmidrule(lr){2-8} 

&
\multicolumn{1}{c|}{\footnotesize{\textit{DDS} + $\left(X,Y\right)^r_{trn}$}} 
& Sim\scriptsize{+conv} & DR&$\checkmark$ & 2,535	  & \textbf{97.8\%} &	\textbf{25.1$^{\circ}$}	\\
\cmidrule(lr){1-8}

&\multicolumn{1}{c|}{$\left(X,Y\right)^r_{trn}$} &
        & 	& $\checkmark$& & 95.4\% &	35.0$^{\circ}$\\

\bottomrule
\vspace{-1.5em} 
\end{tabu}
}
\vspace{-1em} 
\end{table}

The table confirms the benefits of rendering realistic data, with the recognition models trained against previous simulation methods~\cite{gschwandtner2011blensor,planche2017depthsynth} performing almost as well as the instances trained with GAN-based domain adaptation techniques~\cite{bousmalis2017unsupervised,lee2020drit++} having access to a large set of relevant real images. In contrast to the latter methods, simulation tools have, therefore, superior generalization capability.
A second interesting observation from the table is the value of online data augmentation (\eg, random distortion, occlusion, \etc)~\cite{zakharov2018keep}, regardless of the quality of synthetic images. It provides a significant accuracy boost on both tasks, virtually and inexpensively increasing the training set size and variability \cf domain randomization theory~\cite{tobin2017domain}. In that regard, \textit{DeceptionNet}~\cite{zakharov2019deceptionnet}, a learning-based domain randomization framework, performs satisfyingly well without the need for real data (though domain knowledge is required to adequately set the 2.5D transforms' hyperparameters).

But overall, results highlight the benefits of combining all these techniques, which \textit{DDS} can do seamlessly thanks to its gradient-based structure.
Off-the-shelf, manually-parameterized \textit{DDS} yields results similar to previous simulation tools when images are not further augmented but rises above all other methods when adding online augmentations. Training \textit{DDS} unsupervisedly along with $T$ further increases the performance, especially when intermittently applying a learned post-processing composed only of two convolutions. Opting for simple post-processing modules to compensate for non-modelled phenomena, we preserve the key role of simulation within \textit{DDS} and, therefore, its generalization capability.
Finally, we can note that, while the instance of $T$ trained with \textit{DDS} still performs slightly worse than the one trained on real annotated images \wrt the classification task, it outperforms it on the pose estimation task. This is likely due to the finer pose distribution in the rendered dataset (47,268 different images covering every angle of the objects) compared to the smaller real dataset. The best performance \wrt both tasks is achieved by combining the information in the real dataset with simulation-based data (\cf penultimate line in Table~\ref{tab:linemod}).

Though computationally more intensive (a matter that can be offset by rendering images offline), our differentiable solution outperforms all other learning-based domain adaptation schemes, with a fraction of the parameters to train (therefore requiring fewer iterations to converge). Moreover, it is out-of-the-box as valuable as other depth simulation methods and outperforms them too when used within supervised or unsupervised training frameworks.

\subsection{Optimization of Scene and Sensor Parameters}

So far, we mostly focused on optimizing the simulation itself (\eg, shadow bias and noise parameters) in order to render more realistic images and improve CNNs training, rather than optimizing the scene or sensor parameters. 
To illustrate \textit{DDS} capability \wrt such use-cases, we developed and performed a toy experiment, presented in Figure~\ref{fig:scene_sensor_optim}.

\noindent
\textbullet~\textit{Protocol.} 
We consider the same scene setup as in Subsection~\ref{sec:exp-realismstudy} but assume that the target surface is tilted \wrt optical plan and only reflects \textit{red} light frequencies, and that the depth sensor relies on a randomly generated dot pattern emitted with pseudo \textit{white} light (mixture of wavelengths). 

\noindent
\textbullet~\textit{Results.} 
First, in Figure~\ref{fig:scene_sensor_optim}.B, we demonstrate how the scene geometry (\ie, the pose of the flat surface here) can be optimized via gradient descent to reduce the standard error of the simulated device (\ie, using the L1 distance between simulated depth maps and ground-truth noiseless ones as loss function). As expected, the surface is rotated back to be parallel to the focal plane, effectively preventing the stretching of the projected pattern and, therefore, block-matching issues (\cf discussion in Subsection~\ref{sec:exp-realismstudy}).
In a second experiment, we consider the scene parameters fixed and instead try optimizing the depth sensor, focusing on its light pattern (\ie, to reduce sensing errors \wrt this kind of scenes, composed of tilted, red surfaces). Figure~\ref{fig:scene_sensor_optim}.C shows how the pattern image is optimized, quickly switching to  red light frequencies, as well as more slowly adopting local patterns less impacted by projection-induced stretching.

We believe these toy examples illustrate the possible applications of simulation-based optimization of scene parameters (\eg, to reduce noise from surroundings when scanning an object) or sensor parameters (\eg, to build a sensor optimized to specific scene conditions).
\section{Conclusion}

In this paper we presented a novel simulation pipeline for structured-light depth sensors, based on custom differentiable rendering and block-matching operations. While directly performing as well as other simulation tools \wrt generating realistic training images for computer-vision applications, our method can also be further optimized and leveraged within a variety of supervised or unsupervised training frameworks, thanks to its end-to-end differentiability. Such gradient-based optimization can compensate for missing simulation parameters or non-modelled phenomena.
Through various studies, we demonstrate the realistic quality of the synthetic depth images that \textit{DDS} generates, and how depth-based recognition methods can greatly benefit from it to improve their end performance on real data, compared to other simulation tools or learning-based schemes used in scarce-data scenarios.
Our results suggest that the proposed differentiable simulation and its standalone components further bridge the gap between real and synthetic depth data distributions, and will prove useful to larger computer-vision pipelines, as a \textit{transformer} function mapping 3D data and realistic 2.5D scans.

%------------------------------------------------------------------------

{\small
\bibliographystyle{ieee_fullname}
\bibliography{egbib}

\begin{thebibliography}{10}\itemsep=-1pt

\bibitem{akenine2019real}
Tomas Akenine-M{\"o}ller, Eric Haines, and Naty Hoffman.
\newblock {\em Real-time rendering}.
\newblock Crc Press, 2019.

\bibitem{aldoma2012point}
Aitor Aldoma, Zoltan-Csaba Marton, Federico Tombari, Walter Wohlkinger,
  Christian Potthast, Bernhard Zeisl, Radu~Bogdan Rusu, Suat Gedikli, and
  Markus Vincze.
\newblock Point cloud library.
\newblock {\em IEEE Robotics \& Automation Magazine}, 1070(9932/12), 2012.

\bibitem{armeni2017joint}
Iro Armeni, Sasha Sax, Amir~R Zamir, and Silvio Savarese.
\newblock Joint 2d-3d-semantic data for indoor scene understanding.
\newblock {\em arXiv preprint arXiv:1702.01105}, 2017.

\bibitem{bousmalis2016unsupervised}
Konstantinos Bousmalis, Nathan Silberman, David Dohan, Dumitru Erhan, and Dilip
  Krishnan.
\newblock Unsupervised pixel-level domain adaptation with generative
  adversarial networks.
\newblock {\em arXiv preprint arXiv:1612.05424}, 2016.

\bibitem{bousmalis2017unsupervised}
Konstantinos Bousmalis, Nathan Silberman, David Dohan, Dumitru Erhan, and Dilip
  Krishnan.
\newblock Unsupervised pixel-level domain adaptation with generative
  adversarial networks.
\newblock In {\em Proceedings of the IEEE conference on computer vision and
  pattern recognition}, pages 3722--3731, 2017.

\bibitem{chang2018pyramid}
Jia-Ren Chang and Yong-Sheng Chen.
\newblock Pyramid stereo matching network.
\newblock In {\em Proceedings of the IEEE Conference on Computer Vision and
  Pattern Recognition}, pages 5410--5418, 2018.

\bibitem{choo2014statistical}
Benjamin Choo, Michael Landau, Michael DeVore, and Peter~A Beling.
\newblock Statistical analysis-based error models for the microsoft kinecttm
  depth sensor.
\newblock {\em Sensors}, 14(9):17430--17450, 2014.

\bibitem{LearnOpenGLShadowMapping}
Joey de Vries.
\newblock {LearnOpenGL} - {Shadow} {Mapping}.
\newblock
  \href{https://learnopengl.com/Advanced-Lighting/Shadows/Shadow-Mapping}{https://learnopengl.com/Advanced-Lighting/Shadows/Shadow-Mapping}.
  Accessed: 2021-03-10.

\bibitem{denninger2019blenderproc}
Maximilian Denninger, Martin Sundermeyer, Dominik Winkelbauer, Youssef Zidan,
  Dmitry Olefir, Mohamad Elbadrawy, Ahsan Lodhi, and Harinandan Katam.
\newblock Blenderproc.
\newblock {\em arXiv preprint arXiv:1911.01911}, 2019.

\bibitem{drozdzal2016importance}
Michal Drozdzal, Eugene Vorontsov, Gabriel Chartrand, Samuel Kadoury, and Chris
  Pal.
\newblock The importance of skip connections in biomedical image segmentation.
\newblock In {\em Deep learning and data labeling for medical applications},
  pages 179--187. Springer, 2016.

\bibitem{duggal2019deeppruner}
Shivam Duggal, Shenlong Wang, Wei-Chiu Ma, Rui Hu, and Raquel Urtasun.
\newblock Deeppruner: Learning efficient stereo matching via differentiable
  patchmatch.
\newblock In {\em Proceedings of the IEEE/CVF International Conference on
  Computer Vision}, pages 4384--4393, 2019.

\bibitem{einecke2015multi}
Nils Einecke and Julian Eggert.
\newblock A multi-block-matching approach for stereo.
\newblock In {\em 2015 IEEE Intelligent Vehicles Symposium (IV)}, pages
  585--592. IEEE, 2015.

\bibitem{fabius2015variational}
Otto Fabius, Joost~R van Amersfoort, and Diederik~P Kingma.
\newblock Variational recurrent auto-encoders.
\newblock In {\em ICLR (Workshop)}, 2015.

\bibitem{fallon2012efficient}
Maurice~F Fallon, Hordur Johannsson, and John~J Leonard.
\newblock Point cloud simulation \& applications, 2012.
\newblock
  \href{http://www.pointclouds.org/assets/icra2012/localization.pdf}{http://www.pointclouds.org/assets/icra2012/localization.pdf}.
  Accessed: 2020-09-23.

\bibitem{nips12:d3d}
Sanja Fidler, Sven Dickinson, and Raquel Urtasun.
\newblock 3d object detection and viewpoint estimation with a deformable 3d
  cuboid model.
\newblock In {\em Adv. Neural Inform. Process. Syst.}, pages 611--619, 2012.

\bibitem{ganin2015unsupervised}
Yaroslav Ganin and Victor Lempitsky.
\newblock Unsupervised domain adaptation by backpropagation.
\newblock In {\em International Conference on Machine Learning}, pages
  1180--1189, 2015.

\bibitem{ganin2016domain}
Yaroslav Ganin, Evgeniya Ustinova, Hana Ajakan, Pascal Germain, Hugo
  Larochelle, Fran{\c{c}}ois Laviolette, Mario Marchand, and Victor Lempitsky.
\newblock Domain-adversarial training of neural networks.
\newblock {\em The Journal of Machine Learning Research}, 17(1):2096--2030,
  2016.

\bibitem{goodfellow2014generative}
Ian Goodfellow, Jean Pouget-Abadie, Mehdi Mirza, Bing Xu, David Warde-Farley,
  Sherjil Ozair, Aaron Courville, and Yoshua Bengio.
\newblock Generative adversarial nets.
\newblock In {\em NIPS}, pages 2672--2680, 2014.

\bibitem{gschwandtner2011blensor}
Michael Gschwandtner, Roland Kwitt, Andreas Uhl, and Wolfgang Pree.
\newblock Blensor: blender sensor simulation toolbox.
\newblock In {\em Advances in Visual Computing}, pages 199--208. Springer,
  2011.

\bibitem{he2016deep}
Kaiming He, Xiangyu Zhang, Shaoqing Ren, and Jian Sun.
\newblock Deep residual learning for image recognition.
\newblock In {\em CVPR}, pages 770--778, 2016.

\bibitem{hinterstoisser2012model}
Stefan Hinterstoisser, Vincent Lepetit, Slobodan Ilic, Stefan Holzer, Gary
  Bradski, Kurt Konolige, and Nassir Navab.
\newblock Model based training, detection and pose estimation of texture-less
  3d objects in heavily cluttered scenes.
\newblock In {\em ACCV}. Springer, 2012.

\bibitem{hirschmuller2007evaluation}
Heiko Hirschmuller and Daniel Scharstein.
\newblock Evaluation of cost functions for stereo matching.
\newblock In {\em 2007 IEEE Conference on Computer Vision and Pattern
  Recognition}, pages 1--8. IEEE, 2007.

\bibitem{hodavn2019photorealistic}
Tom{\'a}{\v{s}} Hoda{\v{n}}, Vibhav Vineet, Ran Gal, Emanuel Shalev, Jon
  Hanzelka, Treb Connell, Pedro Urbina, Sudipta~N Sinha, and Brian Guenter.
\newblock Photorealistic image synthesis for object instance detection.
\newblock In {\em 2019 IEEE International Conference on Image Processing
  (ICIP)}, pages 66--70. IEEE, 2019.

\bibitem{huber1992robust}
Peter~J Huber.
\newblock Robust estimation of a location parameter.
\newblock In {\em Breakthroughs in statistics}, pages 492--518. Springer, 1992.

\bibitem{humenberger2010fast}
Martin Humenberger, Christian Zinner, Michael Weber, Wilfried Kubinger, and
  Markus Vincze.
\newblock A fast stereo matching algorithm suitable for embedded real-time
  systems.
\newblock {\em Computer Vision and Image Understanding}, 114(11):1180--1202,
  2010.

\bibitem{jaccard1912distribution}
Paul Jaccard.
\newblock The distribution of the flora in the alpine zone. 1.
\newblock {\em New phytologist}, 11(2):37--50, 1912.

\bibitem{jiao2018look}
Jianbo Jiao, Ying Cao, Yibing Song, and Rynson Lau.
\newblock Look deeper into depth: Monocular depth estimation with semantic
  booster and attention-driven loss.
\newblock In {\em Proceedings of the European conference on computer vision
  (ECCV)}, pages 53--69, 2018.

\bibitem{kato2020differentiable}
Hiroharu Kato, Deniz Beker, Mihai Morariu, Takahiro Ando, Toru Matsuoka, Wadim
  Kehl, and Adrien Gaidon.
\newblock Differentiable rendering: A survey.
\newblock {\em arXiv preprint arXiv:2006.12057}, 2020.

\bibitem{keller2009real}
Maik Keller and Andreas Kolb.
\newblock Real-time simulation of time-of-flight sensors.
\newblock {\em Simulation Modelling Practice and Theory}, 17(5):967--978, 2009.

\bibitem{kendall2017end}
Alex Kendall, Hayk Martirosyan, Saumitro Dasgupta, Peter Henry, Ryan Kennedy,
  Abraham Bachrach, and Adam Bry.
\newblock End-to-end learning of geometry and context for deep stereo
  regression.
\newblock In {\em Proceedings of the IEEE International Conference on Computer
  Vision}, pages 66--75, 2017.

\bibitem{kingma2014adam}
Diederik Kingma and Jimmy Ba.
\newblock Adam: A method for stochastic optimization.
\newblock {\em arXiv preprint arXiv:1412.6980}, 2014.

\bibitem{konolige1998small}
Kurt Konolige.
\newblock Small vision systems: Hardware and implementation.
\newblock In {\em Robotics Research}, pages 203--212. Springer, 1998.

\bibitem{konolige2010projected}
Kurt Konolige.
\newblock Projected texture stereo.
\newblock In {\em 2010 IEEE International Conference on Robotics and
  Automation}, pages 148--155. IEEE, 2010.

\bibitem{landau2016}
Michael~J Landau.
\newblock {\em Optimal 6D Object Pose Estimation with Commodity Depth Sensors}.
\newblock PhD thesis, University of Virginia, 2016.
\newblock
  \href{http://search.lib.virginia.edu/catalog/hq37vn57m}{http://search.lib.virginia.edu/catalog/hq37vn57m}.
  Accessed: 2020-10-20.

\bibitem{landau2015simulating}
Michael~J Landau, Benjamin~Y Choo, and Peter~A Beling.
\newblock Simulating kinect infrared and depth images.
\newblock {\em IEEE transactions on cybernetics}, 46(12):3018--3031, 2015.

\bibitem{lee2020drit++}
Hsin-Ying Lee, Hung-Yu Tseng, Qi Mao, Jia-Bin Huang, Yu-Ding Lu, Maneesh Singh,
  and Ming-Hsuan Yang.
\newblock Drit++: Diverse image-to-image translation via disentangled
  representations.
\newblock {\em International Journal of Computer Vision}, 128(10):2402--2417,
  2020.

\bibitem{redner}
Tzu-Mao Li.
\newblock Github - redner: Differentiable rendering without approximation.
\newblock \url{https://github.com/BachiLi/redner}, 2019.
\newblock Accessed: 2021-03-16.

\bibitem{li2018differentiable}
Tzu-Mao Li, Miika Aittala, Fr{\'e}do Durand, and Jaakko Lehtinen.
\newblock Differentiable monte carlo ray tracing through edge sampling.
\newblock {\em ACM Transactions on Graphics (TOG)}, 37(6):1--11, 2018.

\bibitem{iccv13:fpe}
Jasmine~J Lim, Hamed Pirsiavash, and Antonio Torralba.
\newblock Parsing ikea objects: Fine pose estimation.
\newblock In {\em Int. Conf. Comput. Vis.}, pages 2992--2999. IEEE, 2013.

\bibitem{loper2014opendr}
Matthew~M Loper and Michael~J Black.
\newblock Opendr: An approximate differentiable renderer.
\newblock In {\em European Conference on Computer Vision}, pages 154--169.
  Springer, 2014.

\bibitem{luo2016efficient}
Wenjie Luo, Alexander~G Schwing, and Raquel Urtasun.
\newblock Efficient deep learning for stereo matching.
\newblock In {\em Proceedings of the IEEE conference on computer vision and
  pattern recognition}, pages 5695--5703, 2016.

\bibitem{makhzani2015adversarial}
Alireza Makhzani, Jonathon Shlens, Navdeep Jaitly, Ian Goodfellow, and Brendan
  Frey.
\newblock Adversarial autoencoders.
\newblock {\em arXiv preprint arXiv:1511.05644}, 2015.

\bibitem{menna2011geometric}
Fabio Menna, Fabio Remondino, Roberto Battisti, and Erica Nocerino.
\newblock Geometric investigation of a gaming active device.
\newblock In {\em SPIE Optical Metrology}, pages 80850G--80850G. International
  Society for Optics and Photonics, 2011.

\bibitem{michael2013real}
Matthias Michael, Jan Salmen, Johannes Stallkamp, and Marc Schlipsing.
\newblock Real-time stereo vision: Optimizing semi-global matching.
\newblock In {\em 2013 IEEE Intelligent Vehicles Symposium (IV)}, pages
  1197--1202. IEEE, 2013.

\bibitem{nguyen2012modeling}
Chuong~V Nguyen, Shahram Izadi, and David Lovell.
\newblock Modeling kinect sensor noise for improved 3d reconstruction and
  tracking.
\newblock In {\em 3D Imaging, Modeling, Processing, Visualization and
  Transmission (3DIMPVT), 2012 Second International Conference on}, pages
  524--530. IEEE, 2012.

\bibitem{nicodemus1965directional}
Fred~E Nicodemus.
\newblock Directional reflectance and emissivity of an opaque surface.
\newblock {\em Applied optics}, 4(7):767--775, 1965.

\bibitem{kinect3dmodel}
Pierre~Yves P.
\newblock Kinect sensor - 3d warehouse, 2014.
\newblock
  \url{https://3dwarehouse.sketchup.com/model/32ab2192d875d85e58aeac7d536d442b/Kinect-sensor}.
  Accessed: 2021-03-17.

\bibitem{paszke2017automatic}
Adam Paszke, Sam Gross, Soumith Chintala, Gregory Chanan, Edward Yang, Zachary
  DeVito, Zeming Lin, Alban Desmaison, Luca Antiga, and Adam Lerer.
\newblock Automatic differentiation in pytorch.
\newblock 2017.

\bibitem{pharr2016physically}
Matt Pharr, Wenzel Jakob, and Greg Humphreys.
\newblock {\em Physically based rendering: From theory to implementation}.
\newblock Morgan Kaufmann, 2016.

\bibitem{planche2020bridging}
Benjamin Planche.
\newblock {\em Bridging the Realism Gap for CAD-Based Visual Recognition}.
\newblock PhD thesis, University of Passau, 2020.

\bibitem{planche2017depthsynth}
Benjamin Planche, Ziyan Wu, Kai Ma, Shanhui Sun, Stefan Kluckner, Terrence
  Chen, Andreas Hutter, Sergey Zakharov, Harald Kosch, and Jan Ernst.
\newblock Depthsynth: Real-time realistic synthetic data generation from cad
  models for 2.5 d recognition.
\newblock In {\em 3DV}. IEEE, 2017.

\bibitem{planche2019seeing}
Benjamin Planche, Sergey Zakharov, Ziyan Wu, Andreas Hutter, Harald Kosch, and
  Slobodan Ilic.
\newblock Seeing beyond appearance-mapping real images into geometrical domains
  for unsupervised cad-based recognition.
\newblock In {\em 2019 IEEE/RSJ International Conference on Intelligent Robots
  and Systems (IROS)}, pages 2579--2586. IEEE, 2019.

\bibitem{rahman2016optimizing}
Md~Atiqur Rahman and Yang Wang.
\newblock Optimizing intersection-over-union in deep neural networks for image
  segmentation.
\newblock In {\em International symposium on visual computing}, pages 234--244.
  Springer, 2016.

\bibitem{reichinger2011pattern}
A. Reichinger.
\newblock Kinect pattern uncovered.
\newblock \url{http://azttm.wordpress.com/2011/04/03}, 2011.
\newblock Accessed: 2020-03-16.

\bibitem{reitmann2021blainder}
Stefan Reitmann, Lorenzo Neumann, and Bernhard Jung.
\newblock Blainder—a blender ai add-on for generation of semantically labeled
  depth-sensing data.
\newblock {\em Sensors}, 21(6):2144, 2021.

\bibitem{scharstein2007learning}
Daniel Scharstein and Chris Pal.
\newblock Learning conditional random fields for stereo.
\newblock In {\em 2007 IEEE Conference on Computer Vision and Pattern
  Recognition}, pages 1--8. IEEE, 2007.

\bibitem{scharstein2002taxonomy}
Daniel Scharstein and Richard Szeliski.
\newblock A taxonomy and evaluation of dense two-frame stereo correspondence
  algorithms.
\newblock {\em International journal of computer vision}, 47(1):7--42, 2002.

\bibitem{schlick1994inexpensive}
Christophe Schlick.
\newblock An inexpensive brdf model for physically-based rendering.
\newblock In {\em Computer graphics forum}, volume~13, pages 233--246. Wiley
  Online Library, 1994.

\bibitem{nips12:crdl}
Richard Socher, Brody Huval, Bharath Bath, Christopher~D Manning, and Andrew~Y
  Ng.
\newblock Convolutional-recursive deep learning for 3d object classification.
\newblock In {\em Adv. Neural Inform. Process. Syst.}, pages 665--673, 2012.

\bibitem{strasser1974schnelle}
Wolfgang Stra{\ss}er.
\newblock {\em Schnelle kurven-und fl{\"a}chendarstellung auf grafischen
  sichtger{\"a}ten}.
\newblock PhD thesis, 1974.

\bibitem{tobin2017domain}
Josh Tobin, Rachel Fong, Alex Ray, Jonas Schneider, Wojciech Zaremba, and
  Pieter Abbeel.
\newblock Domain randomization for transferring deep neural networks from
  simulation to the real world.
\newblock In {\em IROS}, pages 23--30. IEEE, 2017.

\bibitem{tzeng2017adversarial}
Eric Tzeng, Judy Hoffman, Kate Saenko, and Trevor Darrell.
\newblock Adversarial discriminative domain adaptation.
\newblock {\em arXiv preprint arXiv:1702.05464}, 2017.

\bibitem{tzeng2014deep}
Eric Tzeng, Judy Hoffman, Ning Zhang, Kate Saenko, and Trevor Darrell.
\newblock Deep domain confusion: Maximizing for domain invariance.
\newblock {\em arXiv preprint arXiv:1412.3474}, 2014.

\bibitem{wang2018understanding}
Panqu Wang, Pengfei Chen, Ye Yuan, Ding Liu, Zehua Huang, Xiaodi Hou, and
  Garrison Cottrell.
\newblock Understanding convolution for semantic segmentation.
\newblock In {\em 2018 IEEE winter conference on applications of computer
  vision (WACV)}, pages 1451--1460. IEEE, 2018.

\bibitem{williams1978casting}
Lance Williams.
\newblock Casting curved shadows on curved surfaces.
\newblock In {\em Proceedings of the 5th annual conference on Computer graphics
  and interactive techniques}, pages 270--274, 1978.

\bibitem{Wohlhart15}
Paul Wohlhart and Vincent Lepetit.
\newblock Learning descriptors for object recognition and 3d pose estimation.
\newblock In {\em CVPR}, pages 3109--3118, 2015.

\bibitem{wu2019wider}
Zifeng Wu, Chunhua Shen, and Anton Van Den~Hengel.
\newblock Wider or deeper: Revisiting the resnet model for visual recognition.
\newblock {\em Pattern Recognition}, 90:119--133, 2019.

\bibitem{yu2019simgan}
Simiao Yu, Hao Dong, Felix Liang, Yuanhan Mo, Chao Wu, and Yike Guo.
\newblock Simgan: Photo-realistic semantic image manipulation using generative
  adversarial networks.
\newblock In {\em 2019 IEEE International Conference on Image Processing
  (ICIP)}, pages 734--738. IEEE, 2019.

\bibitem{zakharov2019deceptionnet}
Sergey Zakharov, Wadim Kehl, and Slobodan Ilic.
\newblock Deceptionnet: Network-driven domain randomization.
\newblock In {\em Proceedings of the IEEE/CVF International Conference on
  Computer Vision}, pages 532--541, 2019.

\bibitem{zakharov2017}
Sergey Zakharov, Wadim Kehl, Benjamin Planche, Andreas Hutter, and Slobodan
  Ilic.
\newblock 3d object instance recognition \& pose estimation using triplet loss
  with dynamic margin.
\newblock In {\em IROS}, 2017.

\bibitem{zakharov2018keep}
Sergey Zakharov, Benjamin Planche, Ziyan Wu, Andreas Hutter, Harald Kosch, and
  Slobodan Ilic.
\newblock Keep it unreal: Bridging the realism gap for 2.5 d recognition with
  geometry priors only.
\newblock pages 1--11, 2018.

\bibitem{zhao2020physics}
Shuang Zhao, Wenzel Jakob, and Tzu-Mao Li.
\newblock Physics-based differentiable rendering: from theory to
  implementation.
\newblock In {\em ACM SIGGRAPH 2020 Courses}, pages 1--30. 2020.

\end{thebibliography}
}

%%%%%%%%% SUP MAT
\appendix
\beginsupplement

\newpage
\noindent
\pdfbookmark[0]{Supplementary Material}{sup_mat}
{\Large\textbf{Supplementary Material} \vspace{1em}}

In this supplementary material, we provide further implementation details for reproducibility, as well as additional qualitative and quantitative results.

\begin{figure*}[t]
\centering
\includegraphics[width=1\linewidth]{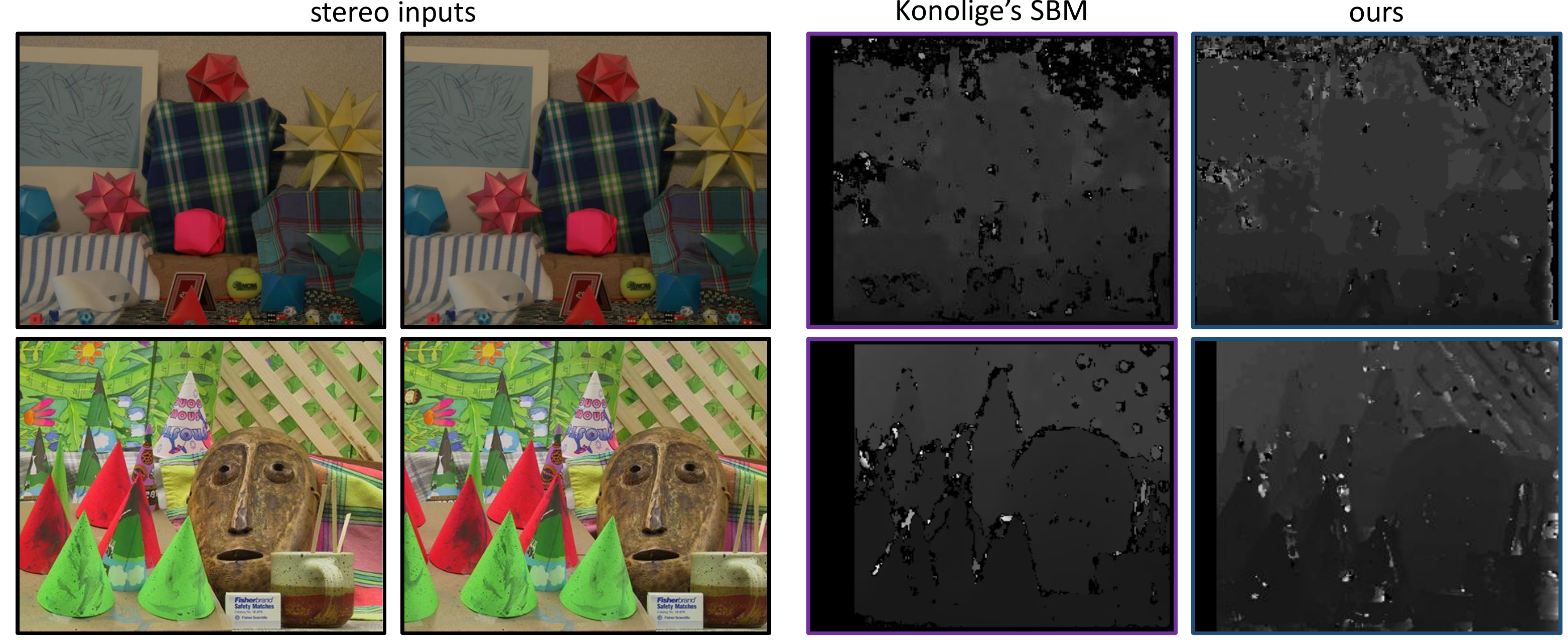}
\vspace{-1.5em}
\caption{\textbf{Comparison of block-matching solutions applied to depth regression from stereo color images}. Our soft block-matching algorithm is compared to Konolige's one~\cite{konolige1998small,konolige2010projected} often used in depth simulation.}
\label{fig:rgb_bm}  
\vspace{-0.5em}
\end{figure*}

\begin{figure*}[t]
\centering
\includegraphics[width=1\linewidth]{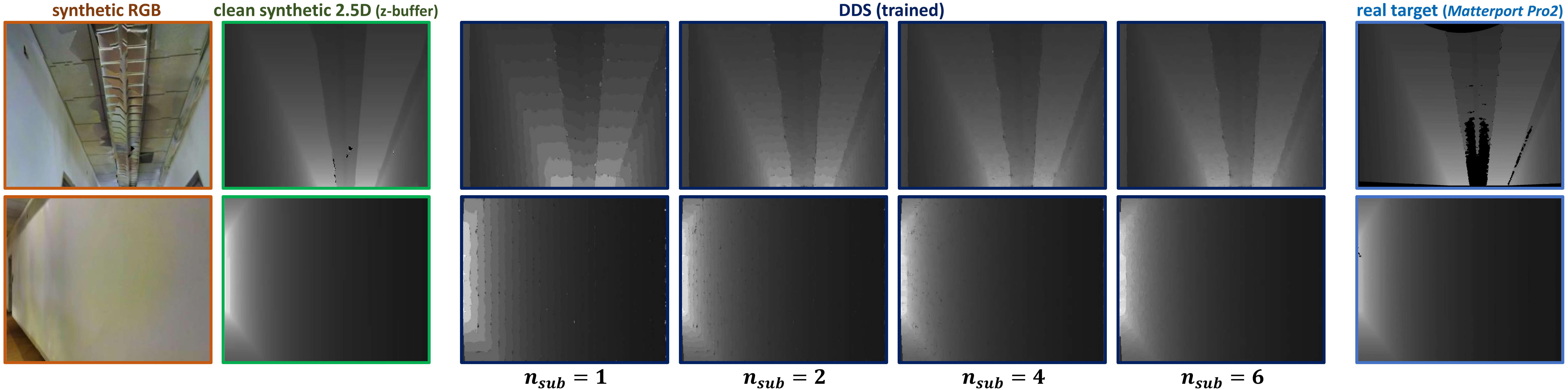}
\vspace{-1.8em}
\caption{\textbf{Impact of proposed differentiable sub-pixel refinement on depth quantization}, depicted over the \textit{2D-3D-Semantic} dataset~\cite{armeni2017joint}.}
\label{fig:subpix_refine}  
\vspace{-0.8em}
\end{figure*}

\begin{figure*}[t]
\centering
\includegraphics[width=1\linewidth]{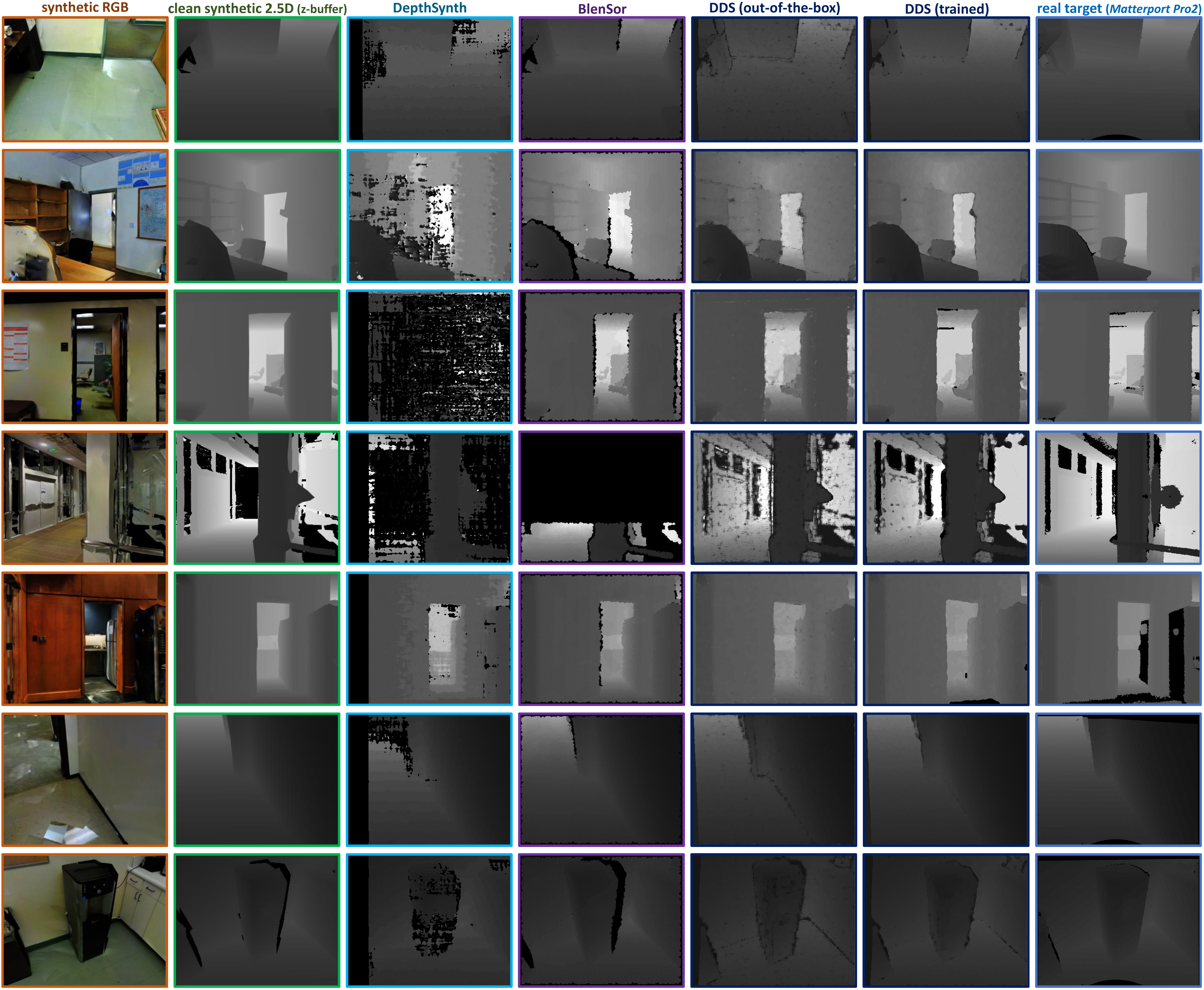}
\vspace{-1.5em}
\caption{\textbf{Qualitative comparison of simulated scans.} Synthetic depth images rendered from reconstructed 3D indoor scenes of the \textit{2D-3D-Semantic} dataset~\cite{armeni2017joint}, compared to real scans from the \textit{Matterport Pro2} sensor. Note that the \textit{Pro2} device relies on 3 stacked depth sensors, hence the high accuracy and reduced shadow noise.}
\label{fig:qual_2d3ds_sup}  
\vspace{-0.8em}
\end{figure*}

\section{Implementation}

\subsection{Practical Details}
Our framework is implemented using PyTorch~\cite{paszke2017automatic}, for seamless integration with optimization and recognition methods. Inference and training procedures are performed on a GPU-enabled backend machine (with two NVIDIA Tesla V100-SXM2 cards).
Differentiable ray-tracing and 3D data processing are performed by the \textit{Redner} tool~\cite{redner} kindly provided by Li~\etal~\cite{li2018differentiable}. 
Optional learning-based post-processing is performed by two convolutional layers, resp. with 32 filters of size $5\times 5$ and 32 filters of size $1\times 1$. The first layer takes as input a 3-channel image composed of the simulated depth map, as well as its noise-free depth map and shadow map (all differentiably rendered by \textit{DDS}).

When optimizing \textit{DDS} (in a supervised or unsupervised manner), we use \textit{Adam}~\cite{kingma2014adam} with a learning rate of $0.001$ and no weight decay. For supervised optimization, we opt for a combination of \textit{Huber} loss~\cite{huber1992robust} and gradient loss~\cite{jiao2018look} (the latter comparing the pseudo-gradient maps obtained from the depth scans by applying \textit{Sobel} filtering).
For unsupervised optimization, we adopt the training scheme and losses from \textit{PixelDA}~\cite{bousmalis2017unsupervised}, \ie, training \textit{DDS} against a discriminator network and in collaboration with the task-specific recognition CNN.

\subsection{Computational Optimization}
On top of the solutions mentioned in the main paper \wrt reducing the computational footprint of \textit{DDS}, we further optimize our pipeline by parallelizing the proposed block-matching algorithm.
Since the correspondence search performed by our method is purely horizontal (\cf horizontal epipolar lines), compared images $I_c$ and $I_o$ can be split into $m$ pairs $\left\{I_{c,j},I_{o,j}\right\}_{j=1}^m$ with:
\begin{equation}\label{eq:parallel}\everymath{\displaystyle}
    I_c = \begin{bmatrix}I_{c,0} \\ I_{c,1} \\ ... \\ I_{c,m}\end{bmatrix} 
    \quad;\quad
    I_o = \begin{bmatrix}I_{o,0} \\ I_{o,1} \\ ... \\ I_{o,m}\end{bmatrix},
\end{equation}
\ie, horizontally splitting the images into $m$ pairs. The stereo block-matching procedure can be performed on each pair independently, enabling computational parallelization (\eg, fixing $m$ as the number of available GPUs).
Note that to account for block size $w \times w$, each horizontal splits $I_{c,j+1}$ and $I_{o,j+1}$ overlaps the previous ones (resp. $I_{c,j}$ and $I_{o,j}$) by $w$ pixels (for notation clarity, Equation~\ref{eq:parallel} does not account for this overlapping).

\subsection{Simulation Parameters}
The results presented in the paper are obtained by providing the following simulation parameters to \textit{DDS} (both as fixed parameters to the off-the-shelf instances and as initial values to the optimized versions):

\paragraph{\textit{Microsoft Kinect V1 Simulation:}}
\begin{itemize}
    \item Image ratio $\frac{H}{W} = \frac{4}{3}$;
    \item Focal length $f_\lambda = 572.41$px;
    \item Baseline distance $b = 75$mm;
    \item Sensor range $[z_{min}, z_{max}] = [400\text{mm},4000\text{mm}]$;
    \item Block size $w = 9$px;
    \item Emitted light intensity factor $\eta_c = 1.5 \times 10^6$;
    \item Shadow bias $\xi = 5$mm;
    \item Softargmax temperature parameter $\beta = 15.0$;
    \item Subpixel refinement level $n_{sub} = 2$;
\end{itemize}

\paragraph{\textit{Matterport Pro2 Simulation:}}
\begin{itemize}
    \item Image ratio $\frac{H}{W} = \frac{5}{4}$;
    \item Focal length $f_\lambda = 1075.43$px;
    \item Baseline distance $b = 75$mm;
    \item Sensor range $[z_{min}, z_{max}] = [400\text{mm},8000\text{mm}]$;
    \item Block size $w = 11$px;
    \item Emitted light intensity factor $\eta_c = 1.5 \times 10^{12}$;
    \item Shadow bias $\xi = 1$mm;
    \item Softargmax temperature parameter $\beta = 25.0$;
    \item Subpixel refinement level $n_{sub} = 4$;
\end{itemize}

Note that device-related parameters come from the sensors' manufacturers or previous \textit{Kinect} studies~\cite{landau2015simulating,landau2016}. Other parameters have been manually set through empirical evaluation. 
For the structured-light pattern, we use the \textit{Kinect} pattern image reverse-engineered by Reichinger~\cite{reichinger2011pattern}.

\begin{figure}[t]
\centering
\includegraphics[width=1\linewidth]{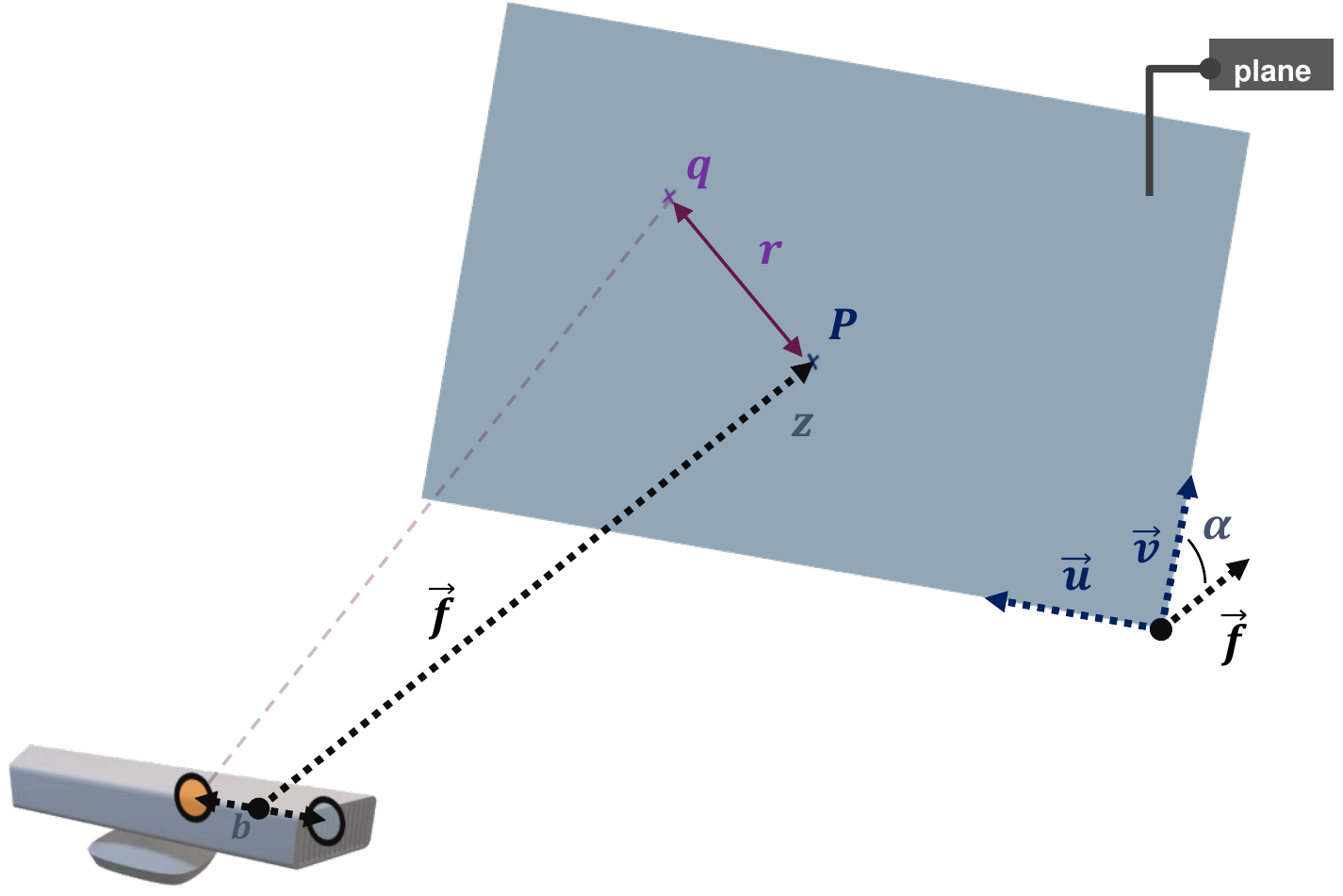}
\vspace{-1.5em}
\caption{\textbf{Experimental setup for quantitative noise study of a depth sensor}, as proposed by Landau~\etal~\cite{landau2015simulating}.}
\label{fig:noisestudy_setup} 
\vspace{-0.8em}
\end{figure}

% \begin{figure*}[t]
% \centering
% \includegraphics[width=1\linewidth]{figures/figure_rebuttal_figonly.pdf}
% \vspace{-1.5em}
% \caption{\textbf{Optimization of scene/sensor parameters} via simulation.}
% \label{fig:scene_sensor_optim} 
% \end{figure*}

\section{Additional Results}

\subsection{Application to RGB Stereo Matching}
Figure~\ref{fig:rgb_bm} provides a glimpse at how the proposed differentiable block-matching algorithm can perform in a stand-alone fashion and be applied to problems beyond the stereo analysis of structured-light patterns. In this figure, our algorithm is applied to the depth measurement of complex stereo color images (without its sub-pixel refinement step, since it relies on ray-tracing). We compare it to the standard stereo block-matching algorithm proposed by Konolige~\cite{konolige1998small,konolige2010projected} and used by previous depth sensor simulations~\cite{gschwandtner2011blensor,planche2017depthsynth}. Stereo color images come from the\textit{ Middlebury Stereo} dataset~\cite{scharstein2002taxonomy,scharstein2007learning,hirschmuller2007evaluation}.
We can appreciate the relative performance of the proposed method, in spite of its excessive quantization (hence the additional sub-pixel refinement proposed in the paper and highlighted in Figure~\ref{fig:subpix_refine}) and approximations for higher-frequency content. We can also observe artifacts for pixels with ambiguous correspondences due to the softargmax-based reduction performed by our method (whereas Konolige's algorithm yields null values when the correspondences are too ambiguous).

\subsection{Realism Study}\label{sec:supmat_realismstudy}
\paragraph{Qualitative Comparison.}
Additional Figure~\ref{fig:subpix_refine} depicts the control over the discrepancy/depth granularity provided by the hyper-parameter $N_{sub}$ (level of subpixel refinement). Incidentally, this figure also shows the impact of non-modelled scene properties on the realism of the simulated scans. The 3D models of the target scenes provided by the dataset authors~\cite{armeni2017joint}, used to render these scans, do not contain texture/material information and have various geometrical defects; hence some discrepancies between the real and synthetic representations (\eg, first row of Figure~\ref{fig:subpix_refine}: the real scan is missing data due to the high reflectivity of some ceiling elements; an information non-modelled in the provided 3D model).
As our pipeline is differentiable not only \wrt the sensor's parameters but also the scene's ones, it could be in theory used to optimize/learn such incorrect or missing scene properties. In practice, this optimization would require careful framing and constraints (worth its own separate study) not to computationally explode , especially for complex, real-life scenes.

Figure~\ref{fig:qual_2d3ds_sup} contains randomly picked synthetic and real images based on the \textit{2D-3D-Semantic} dataset~\cite{armeni2017joint}.
We can observe how the \textit{DepthSynth} method proposed by Planche~\etal~\cite{planche2017depthsynth} tends to over-induce noise, sometimes completely failing at inferring the depth through stereo block-matching. It may be due to the choice of block-matching algorithm~\cite{konolige1998small,konolige2010projected}, as the authors rely on a popular but rather antiquated method, certainly not as robust as the (unspecified) algorithm run by the target \textit{Matterport Pro2} device. Our own block-matching solution is not much more robust (\cf Figure~\ref{fig:rgb_bm}) and also tends to over-induce noise in the resulting depth images. Until a more robust differentiable solution is proposed, \textit{DDS} can, however, rely on its post-processing capability to compensate for the block mismatching and to generate images that are closer to the target ones, as shown in Figure~\ref{fig:qual_2d3ds_sup} (penultimate column).
As for the \textit{BlenSor} simulation~\cite{gschwandtner2011blensor}, its image quality is qualitatively good, though it cannot be configured, \eg, to reduce the shadow noise (the tool proposes a short list of pre-configured sensors that it can simulate). Moreover, for reasons unknown, the open-source version provided by the authors fails to properly render a large number of images from the \textit{2D-3D-S} scenes, resulting in scans missing a large portion of the content (\cf fourth row in Figure~\ref{fig:qual_2d3ds_sup}). This probably explains the low performance of the CNN for semantic segmentation trained over \textit{BlenSor} data. 
Finally, unlike static simulations, the proposed solution can learn to tune down its inherent noise to model more precise sensors such as the multi-shot \textit{Matterport} device (composed of 3 sensors).
 
\paragraph{Quantitative Comparison.}
Figure~\ref{fig:noisestudy_setup} illustrates the experimental setup described in Subsection~\refwithdefault{sec:exp-realismstudy}{4.1} of the paper \wrt noise study. We consider a flat surface placed at distance $z$ from the sensor, with a tilt angle $\alpha$ \wrt the focal plane (with $\overrightarrow{f}$ its normal).

Note that for this experiment, we use the experimental data collected and kindly provided by Landau~\etal~\cite{landau2015simulating}.

\subsection{Applications to Deep Learning}

\begin{table}[t]
\centering
\caption{\textbf{Comparative study \wrt training usage  (extending study in Table~\ref{tab:2d3dds})}, measuring the accuracy of a CNN~\cite{he2016deep,wang2018understanding,wu2019wider} performing semantic segmentation on real 2.5D scans from the indoor \textit{2D-3D-S} dataset~\cite{armeni2017joint}, as a function of the method used to render its training data and as a function of real \textit{annotated} data availability ($\uparrow$ = the higher the value, the better).
}
\label{tab:2d3dds_sup} 
\vspace{-.7em} 
\resizebox{1\linewidth}{!}{
\def\arraystretch{1}% \label{tab:gan_comp}
\begin{tabu}{@{}c|llllllll|c@{}}
\toprule
\multicolumn{1}{c}{\multirow{2}{*}{\shortstack{\\\textbf{Train.}\\\textbf{Data}\\\textbf{Source}}}} & 
\multicolumn{8}{|c}{\textbf{Mean Intersection-Over-Union (mIoU)$^\uparrow$}} & 
\multicolumn{1}{|c}{\multirow{2}{*}{\shortstack{\\[0.2em]\textbf{Pixel}\\\textbf{Acc.$^\uparrow$}}}}
\\
\cmidrule(lr){2-9}
& 

\parbox[t]{1mm}{\rotatebox[origin=c]{45}{\footnotesize\textbf{\textit{bookc.}}}}&
\parbox[t]{2mm}{\rotatebox[origin=c]{45}{\footnotesize\textbf{\textit{ceili.}}}}&
\parbox[t]{2mm}{\rotatebox[origin=c]{45}{\footnotesize\textbf{\textit{chair}}}}&
\parbox[t]{2mm}{\rotatebox[origin=c]{45}{\footnotesize\textbf{\textit{clutter}}}}&
\parbox[t]{2mm}{\rotatebox[origin=c]{45}{\footnotesize\textbf{\textit{door}}}}&
\parbox[t]{2mm}{\rotatebox[origin=c]{45}{\footnotesize\textbf{\textit{floor}}}}&
\parbox[t]{2mm}{\rotatebox[origin=c]{45}{\footnotesize\textbf{\textit{table}}}}&
\parbox[t]{2mm}{\rotatebox[origin=c]{40}{\footnotesize\textbf{\textit{wall}}}}
% & \textbf{mean}  

& \\

\midrule

clean 
& .003 & .018 & .002 & .087 & .012 & .052 & .091 & .351
% & .085
& 35.3\%

\\\cmidrule(lr){1-10}
\textit{BlenSor}{\scriptsize~\cite{gschwandtner2011blensor}} 
& .110 & .534 & .119 & .167 & .148 & .561 & .082 & .412
% & .161
& 51.6\%

\\
\textit{DepthS.}{\scriptsize~\cite{planche2017depthsynth}}
& .184 & .691 & .185 & .221 & .243 & .722 & .235 & .561
% & .264
& 65.3\%

\\
\textit{DDS} 
& .218 & .705 & .201 & .225 & .240 & .742 & .259 & .583
% & .278
& 62.9\%

\\
\textit{DDS} \footnotesize{(train.)} 
& \textbf{.243} & .711 & \textbf{.264} & .255 & .269 & .794 & .271 & .602
% & .296
& 69.8\%

\\\cmidrule(lr){1-10}
% \textit{DDS} + real 
% & XXX & XXX & XXX & XXX & XXX & XXX & XXX & XXX
% % & XXX
% & XXX\%

% \\
real 
& .135 & .770 & .214 & .277 & .302 & .803 & .275 & \textbf{.661}
% & \textbf{.301}
& 73.5\%

\\\cmidrule(lr){1-10}

% \\
\textit{BlenSor}{\scriptsize~\cite{gschwandtner2011blensor}} + real 
& .143 & .769 & .213 & .275 & .306 & \textbf{.817} & .271 & .636 & 73.6\%

\\
\textit{DepthS.}{\scriptsize~\cite{planche2017depthsynth}} + real 
& .222 & .767 & .234 & .297 & .325 & .812 & .273 & .659 & 75.8\%

\\
\textit{DDS} + real 
& \textbf{.279} & \textbf{.775} & \textbf{.245} & \textbf{.299} & \textbf{.356} & .815 & \textbf{.280} & .659 & \textbf{76.7\%}

\\
\bottomrule
\end{tabu}
\vspace{-1em} 
}
%	\vspace{0.4cm}   
\vspace{-1em} 
\end{table}

Table~\ref{tab:2d3dds_sup} extends the results presented in the paper (Table~\refwithdefault{tab:2d3dds}{1}), considering the cases when annotations are provided for the subset of real training images. In such a scenario, the segmentation method can be supervisedly trained either purely on the (rather limited) real data, or on a larger, more varied mix of real and synthetic data. The additional last three rows in Table~\ref{tab:2d3dds_sup} present the test results considering the latter option. We can observe how the CNN instances trained on such larger datasets---and more specifically the CNN instance trained on a mix of real and \textit{DDS} data---are more accurate than the instance trained purely on real data.

Similarly, Table~\ref{tab:linemod_sup} extends the results presented in the paper (Table~\refwithdefault{tab:linemod}{2}) \wrt training of a CNN for instance classification and pose estimation over the \textit{Cropped LineMOD} dataset~\cite{hinterstoisser2012model,bousmalis2017unsupervised,zakharov2019deceptionnet}.
Besides specifying the number of trainable parameters $|\Phi_D|$ that compose discriminator networks (for adversarial domain adaptation methods), we highlight the impact of adding pseudo-realistic clutter to the virtual scenes before rendering images, \ie, adding a flat surface as ground below the target object, and randomly placing additional 3D objects around it.
Intuitive, the benefit of surrounding the target 3D objects with clutter (for single-object image capture) to the realism of the resulting synthetic images has already been highlighted by previous studies on RGB images~\cite{denninger2019blenderproc,hodavn2019photorealistic}.

Our results presented in Table~\ref{tab:linemod_sup} extend these conclusions to the 2.5D domain, with a sharp accuracy increase of the resulting recognition models when adding pseudo-realistic clutter to the virtual scenes. This also highlights the importance, in visual simulation, of not only modeling realistic sensor properties but also of properly setting up the virtual  scenes (\cf discussion in previous Subsection~\ref{sec:supmat_realismstudy}). 
\section{Acknowledgments}
We would like to deeply thank Tzu-Mao Li for the help provided \wrt applying his \textit{Redner} rendering tool~\cite{redner,li2018differentiable} to our needs. 
Finally, credits  go to Pierre Yves P.~\cite{kinect3dmodel} for the 3D \textit{Microsoft Kinect} model used to illustrate some of the figures in our paper.

\begin{table*}[b]
\vspace{3em} 
\centering
\caption{\textbf{Comparative and ablative study (extending study in Table~\refwithdefault{tab:linemod}{2})}, measuring the impact of unsupervised domain adaptation, sensor simulation (Sim), and domain randomization (DR, \ie, using randomized 2.5D transforms to the rendered images \cf \cite{zakharov2018keep,zakharov2019deceptionnet} or adding random 3D clutter to the virtual scenes before rendering) on the training of a CNN~\cite{ganin2015unsupervised} for depth-based instance classification and pose estimation on the \textit{Cropped LineMOD} dataset~\cite{hinterstoisser2012model,bousmalis2017unsupervised,zakharov2019deceptionnet}.
}
\label{tab:linemod_sup} 
\vspace{-.7em} 
\begin{tabu}{@{}cc|c|cc|ccc|cc@{}}
\toprule
\multicolumn{2}{c}{\multirow{2}{*}{}} &
\multicolumn{1}{|c}{\multirow{2}{*}{\shortstack{\textbf{3D Clutter}\\\textbf{in Scene}}}} & 
\multicolumn{2}{|c}{\textbf{Augmentations}} & 
\multicolumn{3}{|c}{\textbf{Sim/DA Req.}} &  
\multicolumn{1}{|c}{\multirow{2}{*}{\shortstack{\textbf{Class.}\\\textbf{Accur.}$^\uparrow$}}} &
\multicolumn{1}{c}{\multirow{2}{*}{\shortstack{\textbf{Rot.}\\\textbf{Error}$^\downarrow$}}}
\\
\cmidrule(lr){4-5}\cmidrule(lr){6-8}
& & & \textbf{offline} & \textbf{online} & $\bm{X^r_{trn}}$ &$\bm{|\Phi|}$ &$\bm{|\Phi_D|}$ &\\

\midrule

&\multicolumn{1}{c|}{\multirow{4}{*}{Basic}} &
&
        &	&	            &	  &    & 21.3\% &	91.8$^{\circ}$
\\
&&
 &      & DR&               &	  &    & 39.6\% &	73.3$^{\circ}$
\\
\cdashlinelr{3-10}
&&
$\checkmark$ &
        &	&	            &	  &    & 46.8\% &	67.0$^{\circ}$
\\
&&
$\checkmark$ &      & DR&               &	  &    & 70.7\% &	53.1$^{\circ}$
\\
\cmidrule(lr){1-10}

\parbox[t]{2mm}{\multirow{10}{*}{\rotatebox[origin=c]{90}{Dom. Adap.}}} &
\multirow{2}{*}{\textit{PixelDA}}{\footnotesize~\cite{bousmalis2017unsupervised}}   & 
&
        & GAN& $\checkmark$	& 1.96M & 693k &	65.8\% &	56.5$^{\circ}$	
\\
\cdashlinelr{3-10}
&&$\checkmark$ &
        & GAN& $\checkmark$	& 1.96M & 693k &	85.7\% &	40.5$^{\circ}$	
\\
\cmidrule(lr){2-10}

& 
\multirow{4}{*}{\textit{DRIT++}{\footnotesize~\cite{lee2020drit++}}}
& 
&   GAN &   & $\checkmark$  & 12.3M & 33.1M&  36.2\% &	91.9$^{\circ}$	\\

&&  & GAN & DR & $\checkmark$  & 12.3M & 33.1M &   62.5\% &	89.1$^{\circ}$	\\
\cdashlinelr{3-10}
&& $\checkmark$
&   GAN &   & $\checkmark$  & 12.3M & 33.1M&  68.0\% &	60.8$^{\circ}$	\\

&&  $\checkmark$ & GAN & DR & $\checkmark$  & 12.3M & 33.1M &   87.7\% &	39.8$^{\circ}$	\\
\cmidrule(lr){2-10}

&
\multirow{2}{*}{\textit{DeceptionNet}{\footnotesize~\cite{zakharov2019deceptionnet}}} & 
&
        & DR & 				& 1.54M & &	37.3\% &	59.8$^{\circ}$	
\\
\cdashlinelr{3-10}
&& $\checkmark$ &
        & DR & 				& 1.54M & &	80.2\% &	54.1$^{\circ}$	
\\
\cmidrule(lr){1-10}
\parbox[t]{2mm}{\multirow{14}{*}{\rotatebox[origin=c]{90}{Sensor Simulation}}}  &
\multirow{4}{*}{\textit{DepthSynth}{\footnotesize~\cite{planche2017depthsynth}}} &
&
   Sim     & 	&	        &	  &     & 17.1\% &	87.5$^{\circ}$	\\
&& &   Sim     & DR	&	        &	  &     & 45.6\% &	65.4$^{\circ}$	\\
\cdashlinelr{3-10}
&&$\checkmark$ &
   Sim     & 	&	        &	  &     & 71.5\% &	52.1$^{\circ}$	\\
&&$\checkmark$ &   Sim     & DR	&	        &	  &     & 76.6\% &	45.4$^{\circ}$	\\
\cmidrule(lr){2-10}

&
\multirow{4}{*}{\textit{BlenSor}{\footnotesize~\cite{gschwandtner2011blensor}}} &
&
   Sim     & 	&	        &	   &    & 14.9\% &	90.1$^{\circ}$	\\
&& &   Sim     & DR	&	        &	 &      & 45.6\% &	65.3$^{\circ}$	\\
\cdashlinelr{3-10}
&&$\checkmark$ &
   Sim     & 	&	        &	   &    & 67.5\% &	63.4$^{\circ}$	\\
&&$\checkmark$ &   Sim     & DR	&	        &	 &      & 82.6\% &	41.4$^{\circ}$	\\
\cmidrule(lr){2-10}

&
\multirow{4}{*}{\shortstack{\textit{DDS}\\(untrained)}} &
&
   Sim     & 	&	        &	    &   & 15.6\% &	91.6$^{\circ}$	\\
&&&   Sim     & DR	&	        &	  &     & 50.0\% &	68.9$^{\circ}$	\\
\cdashlinelr{3-10}
&&$\checkmark$ &
   Sim     & 	&	        &	    &   & 69.7\% &	67.6$^{\circ}$	\\
&&$\checkmark$ &   Sim     & DR	&	        &	  &     & 89.6\% &	39.7$^{\circ}$	\\
\cmidrule(lr){1-10}

\parbox[t]{2mm}{\multirow{10}{*}{\rotatebox[origin=c]{90}{Combined}}} & 
\multirow{8}{*}{\textit{DDS}} &
&
   Sim     & 	& $\checkmark$ & 4 & 693k	  & 21.3\% &	80.9$^{\circ}$	\\
&& &Sim     & DR	& $\checkmark$ & 4 & 693k	  & 51.6\% &	63.3$^{\circ}$	\\
&& &Sim\scriptsize{+conv} & 	& $\checkmark$ & 2,535 & 693k	  & 22.6\% &	78.7$^{\circ}$	\\
&& &Sim\scriptsize{+conv} & DR&$\checkmark$ & 2,535 & 693k	  & 54.3\% &	60.4$^{\circ}$	\\
\cdashlinelr{3-10}

&& $\checkmark$ &
   Sim     & 	& $\checkmark$ & 4 & 693k	  & 81.2\% &	49.1$^{\circ}$	\\
&& $\checkmark$ &Sim     & DR	& $\checkmark$ & 4 & 693k	  & 90.5\% &	39.4$^{\circ}$	\\
&& $\checkmark$ &Sim\scriptsize{+conv} & 	& $\checkmark$ & 2,535 & 693k	  & 85.5\% &	45.4$^{\circ}$	\\
&& $\checkmark$ &Sim\scriptsize{+conv} & DR&$\checkmark$ & 2,535 & 693k	  & 93.0\% &	31.3$^{\circ}$	\\
\cmidrule(lr){2-10} 

&
\multicolumn{1}{c|}{\textit{DDS} + $\left(X,Y\right)^r_{trn}$} 
& 
$\checkmark$ & Sim\scriptsize{+conv} & DR&$\checkmark$ & 2,535 & 693k	  & \textbf{97.8\%} &	\textbf{25.1$^{\circ}$}	\\
\cmidrule(lr){1-10}

&\multicolumn{1}{c|}{$\left(X,Y\right)^r_{trn}$} &
$\checkmark$ &
        & 	& $\checkmark$& & & 95.4\% &	35.0$^{\circ}$\\

\bottomrule
\end{tabu}
\vspace{2em} 
\end{table*}

\end{document}